\documentclass[letterpaper, 10 pt, conference]{ieeeconf}  

\IEEEoverridecommandlockouts                              

\title{\LARGE \bf
FastDSAC: Enhancing Policy Plasticity via Constrained Exploration \\ for Scalable Humanoid Locomotion
}
\author{
Guanchen Lu$^{1}$,
Yajuan Dun$^{2}$,
Yi Zhou$^{1}$,
Letian Tao$^{3}$,
Jingliang Duan$^{4}$,
Jie Li$^{1,*}$,
and Guofa Li$^{1,*}$%
\thanks{This work was supported by the State Key Laboratory of Intelligent Vehicle Safety Technology under Project No. IVSTSKL-202506.}
\thanks{
$^{1}$Chongqing University, Chongqing 400044, China;
$^{2}$Ministry of Industry and Information Technology, Beijing 100804, China;
$^{3}$Tsinghua University, Beijing 100084, China;
$^{4}$University of Science and Technology Beijing, Beijing 100083, China.
$^{*}$Corresponding author: Jie Li
(email: jieli@cqu.edu.cn) and Guofa Li
(email: liguofa@cqu.edu.cn).
}
\\[0.8em]
{\normalfont
{\fontsize{12}{13}\selectfont Project Page:}\hspace{0.3em}
{\fontsize{10}{12}\selectfont
\href{https://luge66.github.io/fastdsac-web/}
{\mbox{https://luge66.github.io/fastdsac-web/}}
}
}
}

\usepackage[hidelinks]{hyperref}
\usepackage{amsmath}
\usepackage{subfigure}
\usepackage{amsfonts}
\usepackage{graphicx}
\usepackage{algorithm}
\usepackage{algpseudocode}
\usepackage{url}
\usepackage{subcaption}
\usepackage{float}

\begin{document}

\maketitle
\thispagestyle{empty}
\pagestyle{empty}

\begin{abstract}
Scalable reinforcement learning has popularized high-throughput sampling architectures, which significantly compresses the training time for off-policy methods in robotic locomotion. However, the rapid increase of data volume and update frequency undermines the stability of value-based methods and diminishes the plasticity of policy networks. To address these challenges, this work presents FastDSAC, a fast and high-performance variant of the Distributional  Actor-Critic algorithm designed for parallel sampling scenarios. Specifically, we introduce a truncated Gaussian distribution to approximate the learned policy, which effectively excludes out-of-distribution actions that strain target value estimation while keeping necessary stochasticity for exploration. The proposed action constraint functions as an implicit regularization, which counteracts the plasticity loss typically caused by aggressive gradient updates. This preservation of network adaptability enhances sample efficiency, particularly in scenarios with a high update-to-data ratio, and accelerates the early training process. In contrast to prior fast reinforcement learning approaches that rely on discrete value distributions, our method utilizes a continuous Gaussian representation equipped with adaptive variance regulation, which improves value estimation accuracy by sampling confident and informative transitions. Extensive experiments on MuJoCo Playground and HumanoidBench demonstrate that FastDSAC not only stabilizes the overall training process but also achieves superior asymptotic performance and faster convergence compared to state-of-the-art baselines. 
\end{abstract}

\section{Introduction} 
Reinforcement learning (RL) has advanced humanoid locomotion and other high-dimensional continuous-control problems~\cite{peng2018deepmimic, heess2017emergence}, yet learning strong policies often remains expensive and sensitive to training conditions. Large-scale parallel simulation improves wall-clock efficiency by increasing data throughput~\cite{makoviychuk2021isaac,rudin2021walkminutes}. However, with the highly parallel sampling and large-batch updates, off-policy actor-critic training can become more sensitive to the noise in TD errors for critic updates~\cite{li2023pql,chen2021redq}. This sensitivity can destabilize value learning and gradually erode learning plasticity~\cite{lyle2022capacity}. Reduced plasticity has been linked to primacy bias and late-stage performance regressions~\cite{nikishin2022primacy}, motivating mitigation strategies such as plasticity injection~\cite{nikishin2023plasticityinjection}.

A key source of this issue is the inaccuracy of the target Q-value estimates used for critic learning. In off-policy actor-critic methods, target Q-values are estimated using target Q-networks~\cite{mnih2015dqn,lillicrap2016ddpg}. During high-throughput training, large mini-batches increase the number of target actions sampled from the target policy, making extreme and low-probability actions more likely and yielding higher-variance target Q-value estimates and TD errors. This increased variance destabilizes critic learning and can slow policy improvement~\cite{fujimoto2019bcq}.

Prior work suggests that stabilizing target Q-values by regularizing the target action can improve off-policy value learning. Techniques such as target policy smoothing~\cite{fujimoto2018td3} and action-space smoothing~\cite{nachum2018smoothie} reduce sensitivity to value-function approximation errors and can be interpreted as forms of value regularization. More broadly, constraining target actions toward the data distribution can mitigate bootstrapping error in off-policy learning~\cite{kumar2019bear}.

In this paper, we propose \textbf{FastDSAC}, a lightweight recipe for stabilizing high-parallel off-policy training in entropy-regularized actor-critic methods. FastDSAC constrains the target action for target Q-value evaluation and derives the corresponding log-probability under the induced action distribution to compute the entropy term. FastDSAC is evaluated on MuJoCo Playground~\cite{zakka2025mujoco} and HumanoidBench~\cite{sferrazza2403humanoidbench}. Under the same high-throughput setting, FastDSAC improves training stability and helps mitigate plasticity loss under high-throughput training, while maintaining comparable wall-clock efficiency to strong off-policy baselines. 
Additionally, we provide open-source code at \url{https://github.com/luge66/FastDSAC}. 

The main contributions of this work are threefold: 
\begin{enumerate}
    \item To more effectively mitigate the variance of value estimation, we propose constraining the target action to a truncated Gaussian distribution during target construction. This method imposes a mean-centered constraint on the support of the induced action distribution, effectively filtering out rare, extreme samples that destabilize the target Q-values, while preserving mild stochasticity for effective exploration.
    \item Benefiting from the constrained exploratory behavior of the policy, which acts as an implicit regularization, the plasticity loss typically associated with high update-to-data (UTD) ratios can be alleviated to some extent. This mechanism enables the strategic deployment of high UTD ratios during the early training phase to accelerate policy convergence, as validated by empirical results on MuJoCo Playground. 
    \item Compared to FastTD3 and FastSAC, which apply discrete value distributions with fixed intervals, the proposed FastDSAC employs continuous Gaussian modeling with adaptive variance regulation to more consistently improve value estimation accuracy and avoid exploring epistemically uncertain regions, thereby retaining more high-reward transitions during interaction. Experiments on standard Humanoid control benchmarks, including MuJoCo Playground and HumanoidBench, demonstrate that FastDSAC has improved the asymptotic performance and reduced the training time compared with other RL algorithms.
\end{enumerate}

\section{Related Work}

\indent\textbf{Massively parallel simulation.}
GPU-accelerated simulation enables massively parallel rollout on a single workstation, making wall-clock efficiency and rapid iteration first-class objectives in robot reinforcement learning. As a result, both on-policy and off-policy pipelines commonly used in robotics, including PPO~\cite{ppo}, SAC~\cite{haarnoja2018sac}, and TD3~\cite{fujimoto2018td3}, are increasingly evaluated and optimized under high-throughput training regimes.
Recent benchmarks and training platforms further strengthen this trend. MuJoCo Playground~\cite{zakka2025mujoco} streamlines MJX-based training for fast single-GPU experimentation, and HumanoidBench~\cite{sferrazza2403humanoidbench} introduces diverse whole-body tasks with high-dimensional actions that expose scalability bottlenecks.
Earlier work has also investigated how to scale off-policy value learning under massively parallel simulation. Parallel Q-Learning~\cite{li2023pql} studies such scaling with a distributional critic~\cite{bellemare2017distributional}, but relies on asynchronous parallelism. More recently, scalable off-policy recipes such as FastTD3~\cite{seo2025fasttd3} and FastSAC~\cite{seo2025learning} show that strong wall-clock performance can be obtained with simple design choices. Overall, massively parallel simulation encourages large-batch learning and higher update-to-data ratios, which can stress value-learning stability and erode learning plasticity.

\indent\textbf{Training Stability and Plasticity.}
Recent studies show that deep RL training can become less stable over time and may suffer from plasticity loss, where agents become less responsive to new data as training progresses~\cite{nikishin2022primacy,lyle2023plasticity,nikishin2023plasticityinjection}.
One line of work improves critic stability by regularizing target value estimation, for example through target policy smoothing~\cite{fujimoto2018td3}, and by using ensemble-based stabilization under high update-to-data ratios, such as REDQ~\cite{chen2021redq} and DroQ~\cite{c17}.
Another complementary direction refines value estimation with distributional critics, ranging from categorical distributions~\cite{bellemare2017distributional} to Gaussian distributional critics (e.g., DSAC)~\cite{duan2025distributional}, and truncation-based distributional variants such as TQC~\cite{c15}.
In addition, objective-level regularization can reduce policy and value churn, such as CHAIN~\cite{tang2024chain}, and alternative value-function training objectives have also been explored~\cite{farebrother2024}.
However, some of these methods improve stability at the cost of additional computation, larger critic ensembles, or more complex training objectives, which can increase wall-clock time and implementation overhead.

\section{Preliminaries}

\subsection{Notation}
We consider an infinite-horizon discounted Markov decision process (MDP)
$(\mathcal{S},\mathcal{A},p,r,\gamma)$, where $\mathcal{S}$ and $\mathcal{A}$ denote the state
and action spaces, $p:\mathcal{S}\times\mathcal{A}\rightarrow\mathcal{P}(\mathcal{S})$ is the
transition kernel, $r:\mathcal{S}\times\mathcal{A}\rightarrow\mathbb{R}$ is the reward function,
and $\gamma\in(0,1)$ is the discount factor. Given a state-action pair $(s,a)$, the next state is
sampled as $s'\sim p(\cdot\,|\,s,a)$. A stochastic policy is denoted by $\pi_\phi(a\,|\,s)$, and
$a\sim\pi_\phi(\cdot\,|\,s)$ and $a'\sim\pi_\phi(\cdot\,|\,s')$ represent actions sampled at
$s$ and $s'$, respectively. The discounted state-action occupancy measure induced by $\pi_\phi$
is denoted by $\rho_{\pi_\phi}(s,a)$. The action-value and state-value functions are denoted
by $Q^{\pi_\phi}(s,a)$ and $V^{\pi_\phi}(s)$; when clear from context, the superscript
$\pi_\phi$ is omitted.

\subsection{Distributional  Actor-Critic}

Let $Z^\pi(s,a)$ denote the  state-action return random variable under policy $\pi$~\cite{bellemare2017distributional,haarnoja2018sac}.
Its expectation recovers the  action-value function.
Distributional  Policy Iteration models the conditional distribution of $Z^\pi(s,a)$ and defines the distributional  Bellman backup
\begin{equation}
\mathcal{T}^{\pi}_{\mathcal{D}} Z(s,a)
\;{\overset{D}{=}}\;
r+\gamma\Big(Z(s',a')-\alpha\log\pi(a'\,|\,s')\Big),
\label{eq:dspi_backup_short}
\end{equation}
where $\overset{D}{=}$ denotes equality in distribution ~\cite{bellemare2017distributional}.
In practice, the value distribution is updated to match the target distribution induced by \eqref{eq:dspi_backup_short}, commonly via divergence minimization ~\cite{duan2025distributional}.

DSAC builds on this principle with an off-policy actor-critic architecture~\cite{duan2025distributional}. The critic parameterizes a Gaussian value distribution
\begin{equation}
Z_{\theta}(\cdot\,|\,s,a)=\mathcal{N}\!\big(Q_{\theta}(s,a),\,\sigma_{\theta}^2(s,a)\big),
\label{eq:gauss_value_dist_short}
\end{equation}
and we use a slowly updated target network with parameters $\theta'$ to construct stable learning targets.

\paragraph{Policy evaluation}
Given a transition $(s,a,r,s')$ sampled from the replay buffer, we sample
$a' \sim \pi_{\phi}(\cdot\,|\,s')$ and $Z' \sim Z_{\theta'}(\cdot\,|\,s',a')$,
and form the random target return
\begin{equation}
y_z = r + \gamma\big(Z' - \alpha \log \pi_\phi(a'|s')\big).
\end{equation}
The critic is trained by maximizing the likelihood of $y_z$ under $Z_{\theta}(\cdot\,|\,s,a)$~\cite{duan2025distributional}.
Under the Gaussian parameterization in \eqref{eq:gauss_value_dist_short}, the loss becomes
\begin{equation}
J_Z(\theta)
=
\mathbb{E}\!\left[
\frac{\big(y_z-Q_{\theta}(s,a)\big)^2}{2\,\sigma_{\theta}^2(s,a)}
+\log\sigma_{\theta}(s,a)
\right]
+\text{const}.
\label{eq:gauss_nll_short}
\end{equation}

\paragraph{Policy improvement}
The actor is learned by maximizing the  objective using the critic mean value~\cite{haarnoja2018sac}:
\begin{equation}
J_{\pi}(\phi)
=
\underset{\substack{s\sim\mathcal{B}\\ a\sim\pi_{\phi}(\cdot\,|\,s)}}{\mathbb{E}}
\!\left[
Q_{\theta}(s,a)-\alpha\log\pi_{\phi}(a\,|\,s)
\right].
\label{eq:actor_obj_short}
\end{equation}

\section{Method}
Our method is designed for high-parallel off-policy training. In entropy-regularized actor-critic methods, policy exploration is encouraged by policy entropy and the increased diversity of experience from large-scale parallel data collection. However, when scaled up, the target Q-value can become more sensitive to occasional extreme target actions, which can increase the variance of the target Q-value estimate and destabilize critic learning. 

To address this, we decouple exploration from target Q-value estimation and impose regularization exclusively on the target action utilized by the target Q-network when forming the TD error. By constraining the target action during target evaluation, we mitigate the adverse impact of rare, extreme target actions on the target Q-value estimate, thereby yielding lower-variance TD errors, improving training stability, and helping mitigate plasticity loss over training. Furthermore, the entropy term is computed using the log-probability consistent with the constrained target action.

\subsection{Gaussian Policy Parameterization}

We parameterize the actor as a diagonal Gaussian policy. Concretely, given a state $s$, the policy network outputs the mean $\mu_\phi(s)$ and standard deviation $\sigma_\phi(s)$ of the Gaussian distribution, i.e., the stochastic policy is defined as $\pi_\phi(a\,|\,s) \sim \mathcal N(\mu_\phi(s),\sigma_\phi^2(s))$. In order to allow the policy gradient to propagate back through the sampling process during training, the reparameterization trick is employed:
\begin{equation}
a=\mu_\phi(s)+\sigma_\phi(s)\odot\epsilon,\quad \epsilon \sim \mathcal{N}(0,I),
\end{equation}
and the log-probability is computed in closed-form as
\begin{equation}
\log\pi_\phi(a\,|\,s)=\sum_{i=1}^{d}\log \mathcal{N}\!\left(a_i\,|\,\mu_\phi^{\,i}(s),(\sigma_\phi^{\,i}(s))^2\right),
\end{equation}
where $d$ denotes the action dimension.

Compared with the commonly used tanh-squashed parameterization, this design offers two advantages in practical policy optimization.
First, the tanh nonlinearity can cause vanishing gradients when actions approach the bounds, which slows down policy learning.
Second, it avoids the log-likelihood correction induced by tanh squashing, which simplifies implementation and avoids the numerical issues associated with the change-of-variables correction.
For reference, a tanh-squashed policy samples a raw action $u\sim\pi_\phi(u\,|\,s)$ and then maps it to the bounded action via $a=\tanh(u)$. The resulting log-probability takes the form:
\begin{equation}
\log\pi_\phi(a\,|\,s)=\log\pi_\phi(u\,|\,s)-\sum_{i=1}^{d}\log\bigl(1-\tanh^2(u_i)\bigr).
\end{equation}

\begin{algorithm}[t]
\caption{FastDSAC (Distributional critic omitted)}
\label{alg:FastDSAC}
\begin{algorithmic} 
\Statex Initialize critic networks $Q_{\theta_1}, Q_{\theta_2}$ and actor network $\pi_{\phi}$ 
\Statex Initialize target networks $\theta_1' \leftarrow \theta_1,\ \theta_2' \leftarrow \theta_2$
\Statex Initialize replay buffer $\mathcal{B}$

\For{$t = 1$ to $T$}
    \State Select action $a \sim \pi_{\phi}(\cdot|\,s)$
    \State Get reward $r$ and new state $s'$
    \State Store samples $\tau = (s,a,r,s')$ in $\mathcal{B}$
    \For{$j = 1$ to $\mathrm{num\_updates}$}
        \State Sample mini-batch $B=\{\tau_k\}_{k=1}^{|B|}$ from $\mathcal{B}$
        \State Compute target Q-value:
        \State $ a'=\mu_{\phi}(s') + c\,\tanh(u'-\mu_{\phi}(s')),\;u'\sim\pi_{\phi}(\cdot|\,s') $
        \State $y=r+\gamma\big(Q_{\theta'}^{\min}(s',a')- \alpha \log \pi_{\phi}(a'|\,s') \big)$
        
        \State Update critic:
        \State $\theta_i \leftarrow \theta_i - \nabla_{\theta_i}\frac{1}{|B|}\sum_{}\big(Q_{\theta_i}(s,a)-y\big)^2$
        \State \textbf{if} $t \bmod d$ = 0 \textbf{then}
        \State \hspace{0.4em} Sample $\tilde{a}\sim\pi_{\phi}(\cdot\mid s)$ using reparameterization
        \State \hspace{0.4em} Update actor:
        \State \hspace{0.4em} $\phi \leftarrow \phi - \nabla_{\phi}\frac{1}{|B|}\sum_{}\big(\alpha \log \pi_{\phi}(\tilde{a}\, |\,s)-Q_{\theta}^{\min}(s,\tilde{a})\big)$
        \State \hspace{0.4em} Update entropy temperature:
        \State \hspace{0.4em} $
        \alpha \leftarrow \alpha - \nabla_{\alpha}\frac{1}{|B|}\sum_{}\big(\mathcal{H}^{\mathrm{target}}-\mathcal{H}(s)\big)\cdot \alpha$
        \State \textbf{end if}
        \State Update target critic:
        \State  $\theta_i' \leftarrow \rho\,\theta_i' + (1-\rho)\theta_i$
    \EndFor
\EndFor
\end{algorithmic}
\end{algorithm}

In contrast, the Gaussian policy without tanh squashing evaluates $\log\pi_\phi(a\,|\,s)$ directly under the Gaussian distribution. This avoids the change-of-variables correction in common reinforcement learning implementations, simplifying probability evaluation and improving numerical stability.

To further improve numerical stability, the policy mean $\mu_\phi(s)$ is bounded within $[m_{\min},\,m_{\max}]$, which prevents excessively large mean actions in the early training and stabilizes optimization. The policy standard deviation $\sigma_\phi(s)$ is initialized to a small value and constrained within $[\ell_{\min},\,\ell_{\max}]$, which keeps the policy standard deviation well controlled and improves training stability.

\subsection{Mean-Centered Truncation for Target Actions}

This subsection introduces a mean-centered truncation constraint on the target action used to compute the target Q-values and derives the log-probability under the induced action distribution.
Given the next state $s'$, the policy network outputs $\mu_\phi(s')$ and $\sigma_\phi(s')$, which parameterize a diagonal Gaussian.
First, we sample a pre-truncation action
\begin{equation}
u' = \mu_\phi(s') + \sigma_\phi(s') \odot \epsilon,\quad \epsilon\sim\mathcal{N}(0,I),
\end{equation}
and then obtain the target action via a mean-centered truncation mapping
\begin{equation}
a' = \mu_\phi(s') + c\,\tanh\!\big(u' - \mu_\phi(s')\big),
\label{eq:trunc_map}
\end{equation}
the elementwise $\tanh(\cdot)$ smoothly squashes the offset $u'-\mu_\phi(s')$, so each component of $a'-\mu_\phi(s')$ is strictly bounded within $[-c,\,c]$.
Compared with hard clipping, this mean-centered truncation mapping is continuous and avoids boundary discontinuities, leading to smoother target evaluation.
Importantly, we apply this mapping only when evaluating the target Q-value $Q_{\theta'}(s',a')$ at the next state.

Because the target Q-values are evaluated at the mean-centered truncated action $a'$, the entropy term is required to use the log-probability associated with the corresponding induced action distribution.
With $u'\sim\pi_\phi(\cdot\,|\,s')$ sampled from the pre-truncation diagonal-Gaussian, the change-of-variables rule applied to the truncation mapping~\eqref{eq:trunc_map} gives
\begin{equation}
\log \pi_\phi(a'\,|\,s')
= \log \pi_\phi(u'\,|\,s')
-\log\!\left|\det\!\left(\frac{\partial a'}{\partial u'}\right)\right|,
\end{equation}
where $\log \pi_\phi(u'\,|\,s')$ is the diagonal-Gaussian log-probability and the second term is the log absolute Jacobian determinant of the mapping from $u'$ to $a'$.
Since the mapping is applied independently to each action dimension, the Jacobian is diagonal, and the log absolute determinant is given by:
\begin{equation}
\log\!\left|\det\!\left(\frac{\partial a'}{\partial u'}\right)\right|
= \sum_{i=1}^{d} \log\!\left( c\,[1-\tanh^2(u'_i-\mu_\phi^{\,i}(s'))] \right),
\label{eq:target_side_logp}
\end{equation}
where $d$ is the action dimension. 

\begin{figure*}[t!]
    \centering

    \begin{minipage}[t]{0.24\textwidth}
        \centering
        \includegraphics[width=1.05\linewidth,height=3.1cm,keepaspectratio=false]{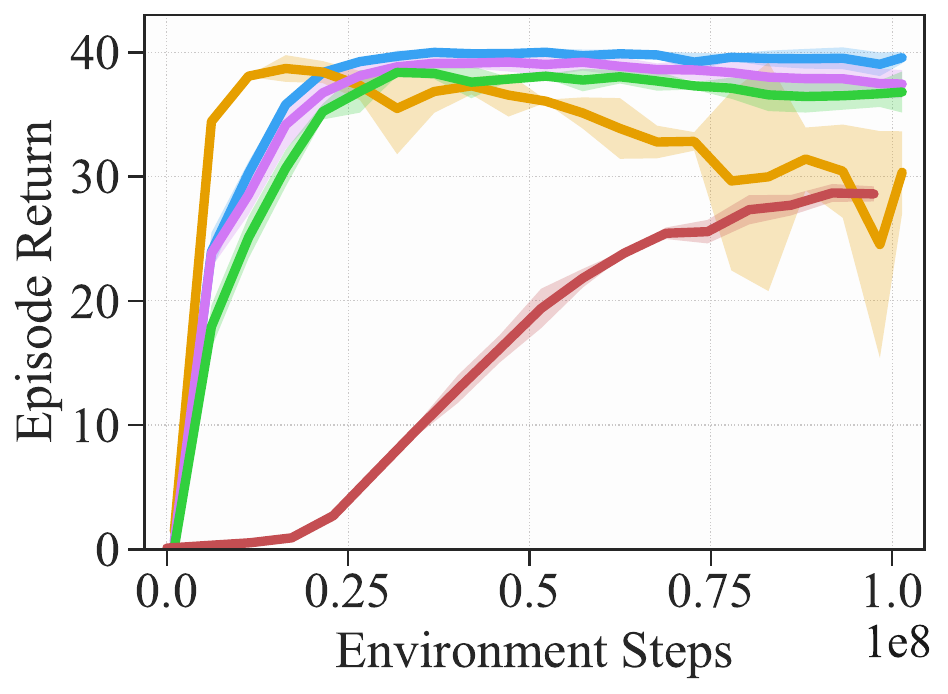}\\
        \vspace{0.01cm}
        \small \hspace{7mm}\textbf{(a)} T1JoystickFlatTerrain\par
    \end{minipage}
    \hspace{0.007\textwidth}
    \begin{minipage}[t]{0.24\textwidth}
        \centering
        \includegraphics[width=\linewidth]{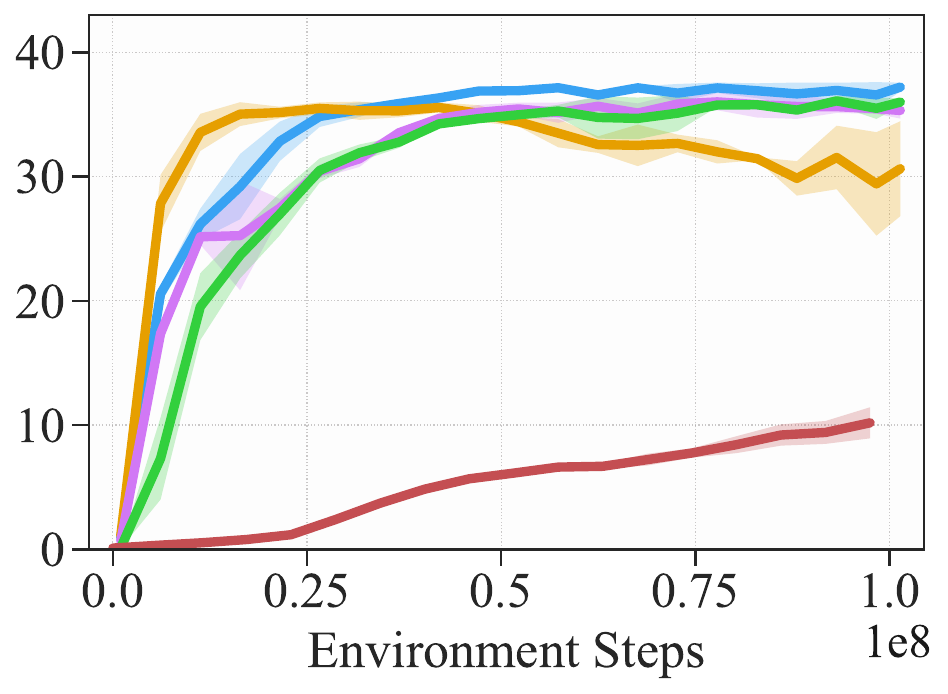}\\
        \vspace{0.01cm}
        \small \hspace{2mm}\textbf{(b)} T1JoystickRoughTerrain\par
    \end{minipage}
    \hspace{-0.005\textwidth}
    \begin{minipage}[t]{0.24\textwidth}
        \centering
        \includegraphics[width=\linewidth]{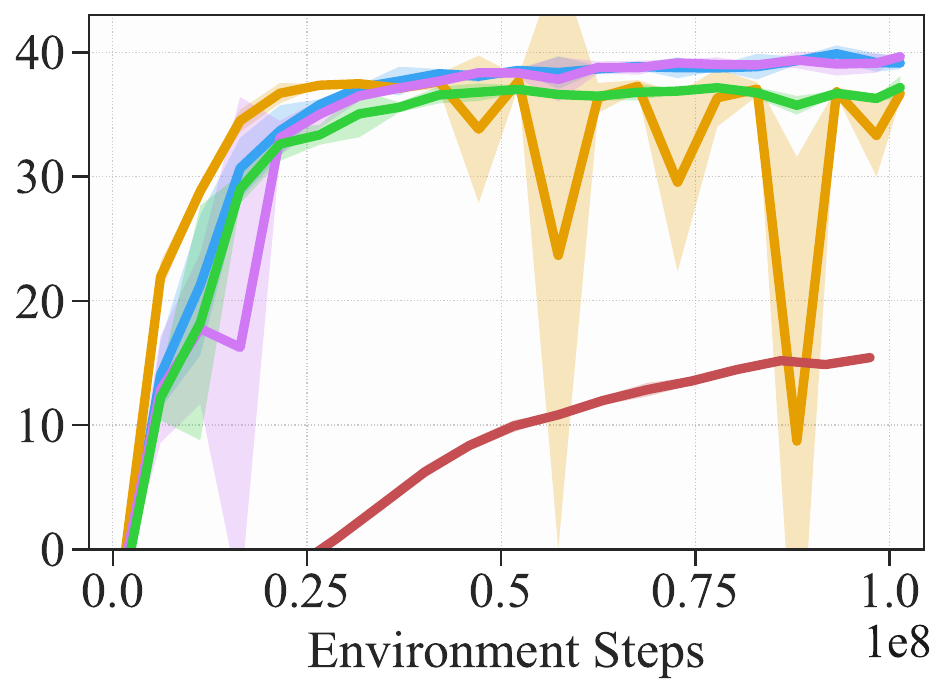}\\
        \vspace{0.01cm}
        \small \hspace{2mm}\textbf{(c)} G1JoystickFlatTerrain\par
    \end{minipage}
    \hspace{-0.005\textwidth}
    \begin{minipage}[t]{0.24\textwidth}
        \centering
        \includegraphics[width=\linewidth]{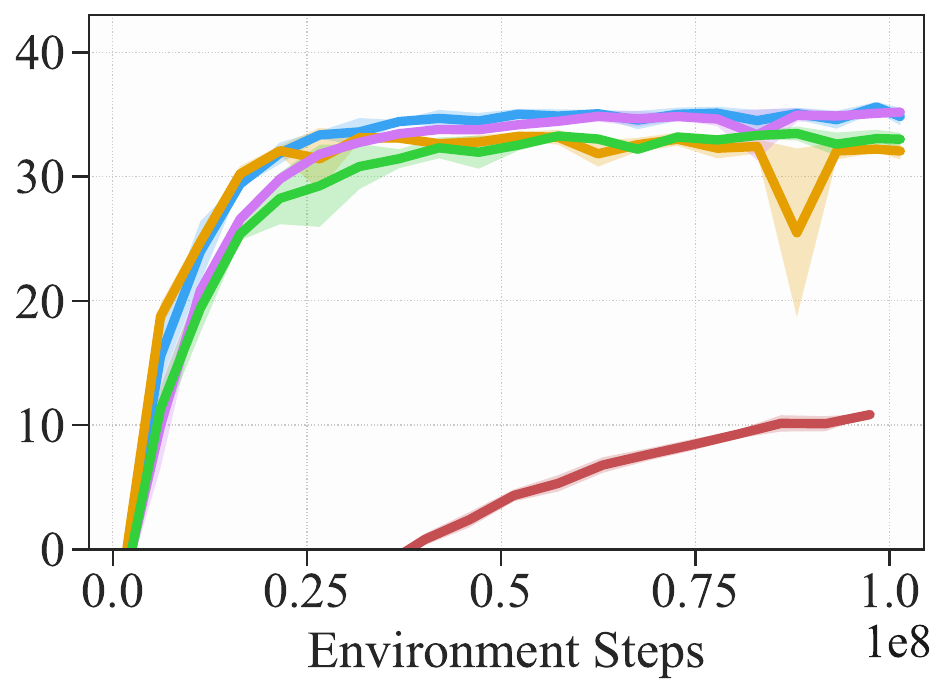}\\
        \vspace{0.01cm}
        \small \hspace{2mm}\textbf{(d)} G1JoystickRoughTerrain\par
    \end{minipage}

    \vspace{0.2cm}

    \begin{minipage}[t]{1.0\textwidth}
        \centering
        \includegraphics[width=\linewidth]{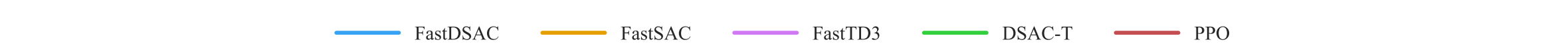}\\
        
    \end{minipage}

    \caption{\textbf{Episode returns versus environment steps on MuJoCo Playground.}
    Four locomotion tasks are shown: T1JoystickFlatTerrain, T1JoystickRoughTerrain, G1JoystickFlatTerrain, and G1JoystickRoughTerrain.
    The x-axis denotes environment steps and the y-axis denotes episode return.
    Solid lines show the mean over three random seeds, and shaded regions indicate one standard deviation.}
    \label{fig:mjstep_results}
\end{figure*}

\begin{figure*}[t!]
    \centering

    \begin{minipage}[t]{0.24\textwidth}
        \centering
        \includegraphics[width=1.05\linewidth,height=3.2cm,keepaspectratio=false]{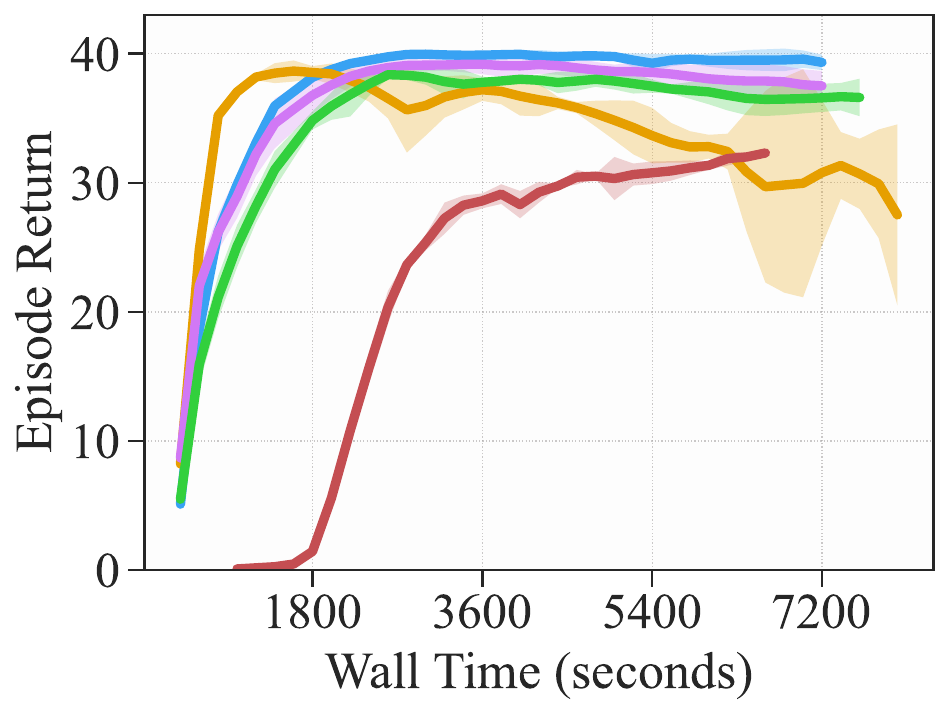}\\
        \vspace{0.01cm}
        \small \hspace{7mm}\textbf{(a)} T1JoystickFlatTerrain\par
    \end{minipage}
    \hspace{0.007\textwidth}
    \begin{minipage}[t]{0.24\textwidth}
        \centering
        \includegraphics[width=\linewidth]{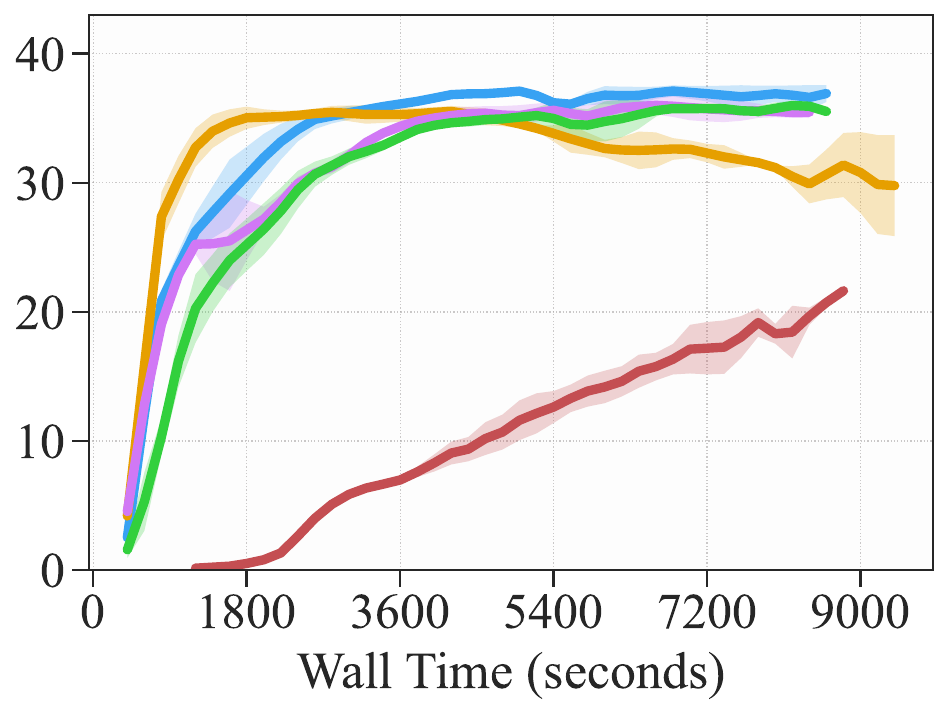}\\
        \vspace{0.01cm}
        \small \hspace{2mm}\textbf{(b)} T1JoystickRoughTerrain\par
    \end{minipage}
    \hspace{-0.005\textwidth}
    \begin{minipage}[t]{0.24\textwidth}
        \centering
        \includegraphics[width=\linewidth]{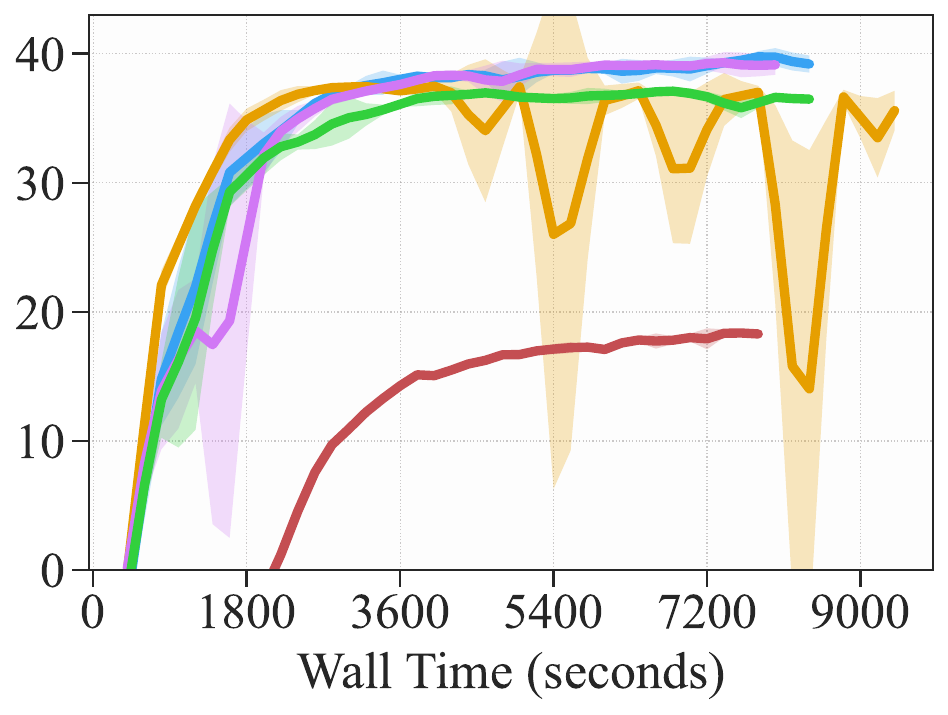}\\
        \vspace{0.01cm}
        \small \hspace{2mm}\textbf{(c)} G1JoystickFlatTerrain\par
    \end{minipage}
    \hspace{-0.005\textwidth}
    \begin{minipage}[t]{0.24\textwidth}
        \centering
        \includegraphics[width=\linewidth]{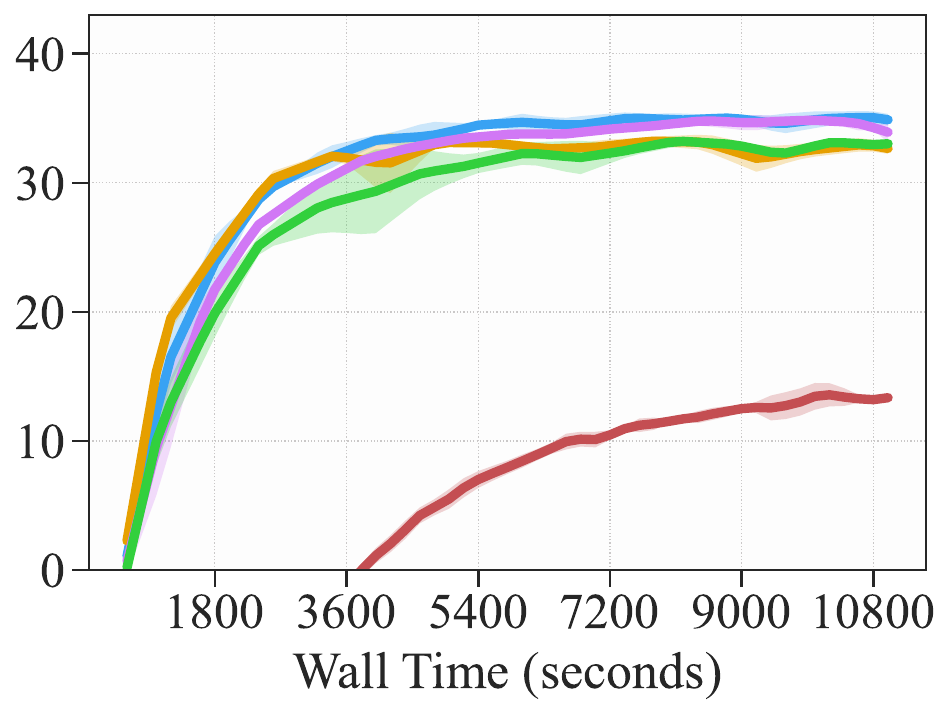}\\
        \vspace{0.01cm}
        \small \hspace{2mm}\textbf{(d)} G1JoystickRoughTerrain\par
    \end{minipage}

    \vspace{0.2cm}

    \begin{minipage}[t]{1.0\textwidth}
        \centering
        \includegraphics[width=\linewidth]{QQ20260306-105634.png}\\
        
    \end{minipage}

    \caption{\textbf{Episode returns versus wall-clock training time on MuJoCo Playground.}
    The same four tasks as in Fig.~\ref{fig:mjstep_results} are shown.
    The x-axis denotes wall-clock training time in seconds and the y-axis denotes episode return.
    Solid lines show the mean over three random seeds, and shaded regions indicate one standard deviation.}
    \label{fig:mjclock_results}
\end{figure*}

For numerical stability in PyTorch implementations, the term $1-\tanh^2(\cdot)$ may suffer from precision loss when $\tanh(\cdot)$ saturates. Accordingly, the Jacobian term is rewritten in an equivalent softplus form:
\begin{equation}
\log\!\bigl(1-\tanh^2(\tilde{u}'_i)\bigr)
= 2\Bigl(\log 2 - \tilde{u}'_i - \mathrm{softplus}(-2\tilde{u}'_i)\Bigr),
\end{equation}
where $\tilde{u}'_i = u'_i-\mu_\phi^{\,i}(s')$ denotes the mean-centered offset. With this identity substituted into the log absolute determinant~\eqref{eq:target_side_logp}, a numerically stable evaluation of the induced log-probability $\log \pi_\phi(a'\,|\,s')$ is obtained in PyTorch. 

The mean-centered truncation directly restricts the target action used to form the TD error, thereby reducing the influence of rare, extreme next actions on the target Q-values.
Moreover, the induced log-probability is made consistent with the constrained action, so the entropy regularizer in the maximum-entropy objective remains properly aligned with the resulting action distribution.
At the same time, the mapping preserves small local perturbations around $\mu_\phi(s')$, retaining the stochasticity needed for learning.
As a result, critic learning becomes less sensitive to extremes, improving training stability and helping alleviate plasticity loss. 

\begin{figure*}[t!]
    \centering

    \begin{minipage}[t]{0.24\textwidth}
        \centering
        \includegraphics[width=1.05\linewidth,height=3.1cm,keepaspectratio=false]{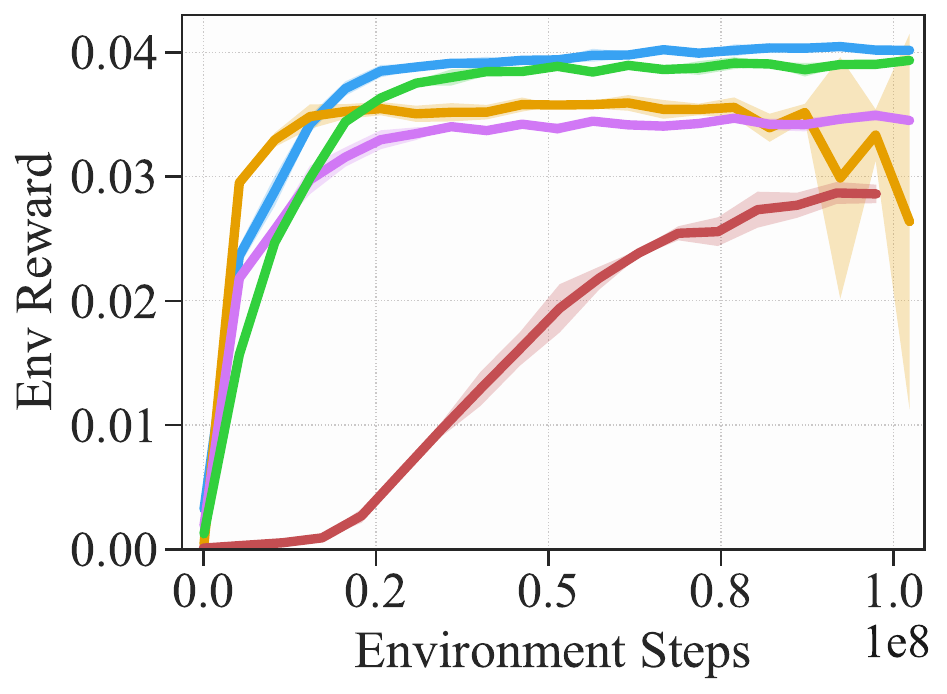}\\
        \vspace{0.01cm}
        \small \hspace{7mm}\textbf{(a)} T1JoystickFlatTerrain\par
    \end{minipage}
    \hspace{0.007\textwidth}
    \begin{minipage}[t]{0.24\textwidth}
        \centering
        \includegraphics[width=\linewidth]{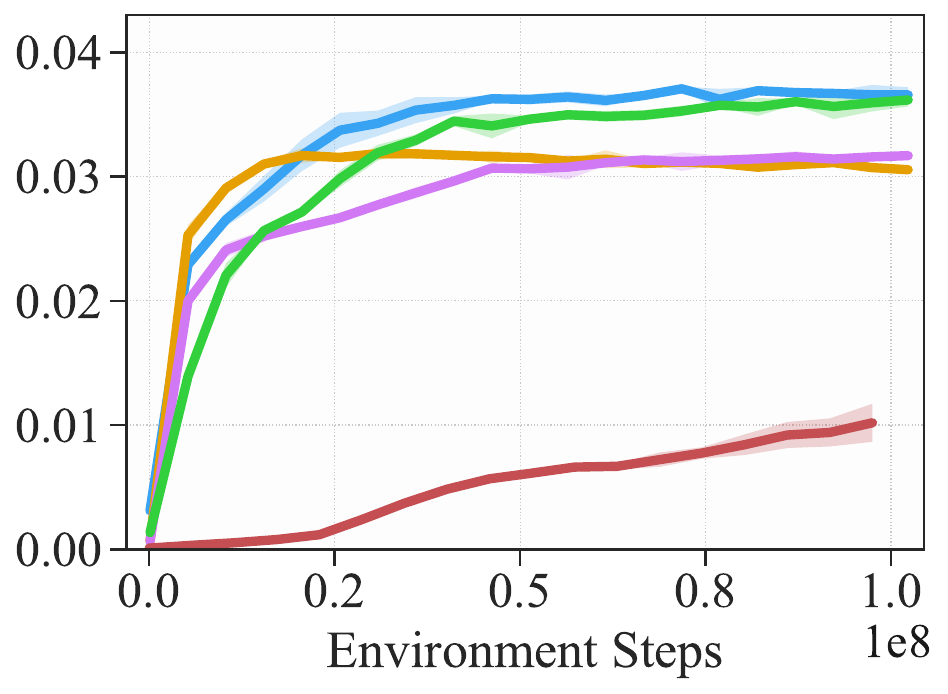}\\
        \vspace{0.01cm}
        \small \hspace{2mm}\textbf{(b)} T1JoystickRoughTerrain\par
    \end{minipage}
    \hspace{-0.005\textwidth}
    \begin{minipage}[t]{0.24\textwidth}
        \centering
        \includegraphics[width=\linewidth]{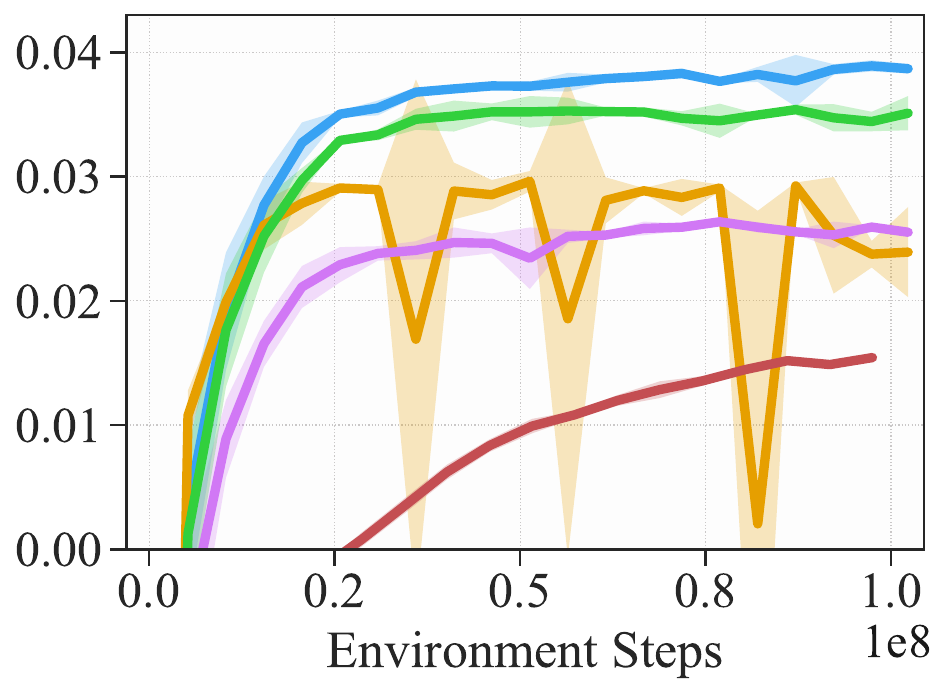}\\
        \vspace{0.01cm}
        \small \hspace{2mm}\textbf{(c)} G1JoystickFlatTerrain\par
    \end{minipage}
    \hspace{-0.005\textwidth}
    \begin{minipage}[t]{0.24\textwidth}
        \centering
        \includegraphics[width=\linewidth]{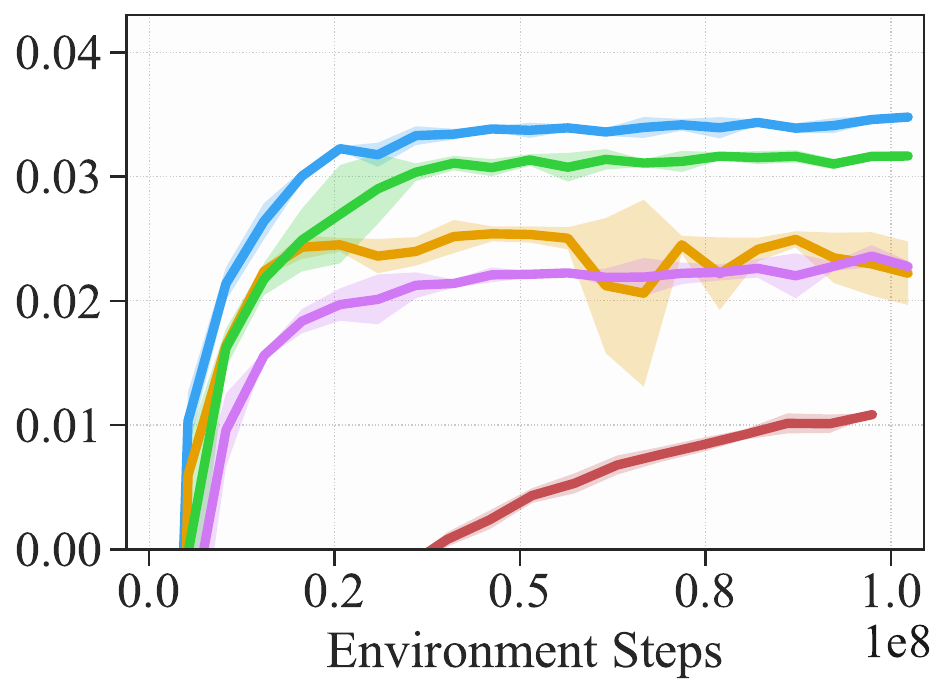}\\
        \vspace{0.01cm}
        \small \hspace{2mm}\textbf{(d)} G1JoystickRoughTerrain\par
    \end{minipage}

    \vspace{0.2cm}

    \begin{minipage}[t]{1.0\textwidth}
        \centering
        \includegraphics[width=\linewidth]{QQ20260306-105634.png}\\
        
    \end{minipage}

    \caption{\textbf{Transition rewards during data collection on MuJoCo Playground.}
    The figure reports the per-transition reward $r$ from collected tuples $\langle s,a,r,s'\rangle$ on the same four tasks.
    The x-axis denotes environment steps and the y-axis denotes transition reward.
    Solid lines show the mean over three random seeds, and shaded regions indicate one standard deviation.}
    \label{fig:mjreward_results}
\end{figure*}

\begin{figure*}[t!]
    \centering

    \begin{minipage}[t]{0.24\textwidth}
        \centering
        \includegraphics[width=\linewidth]{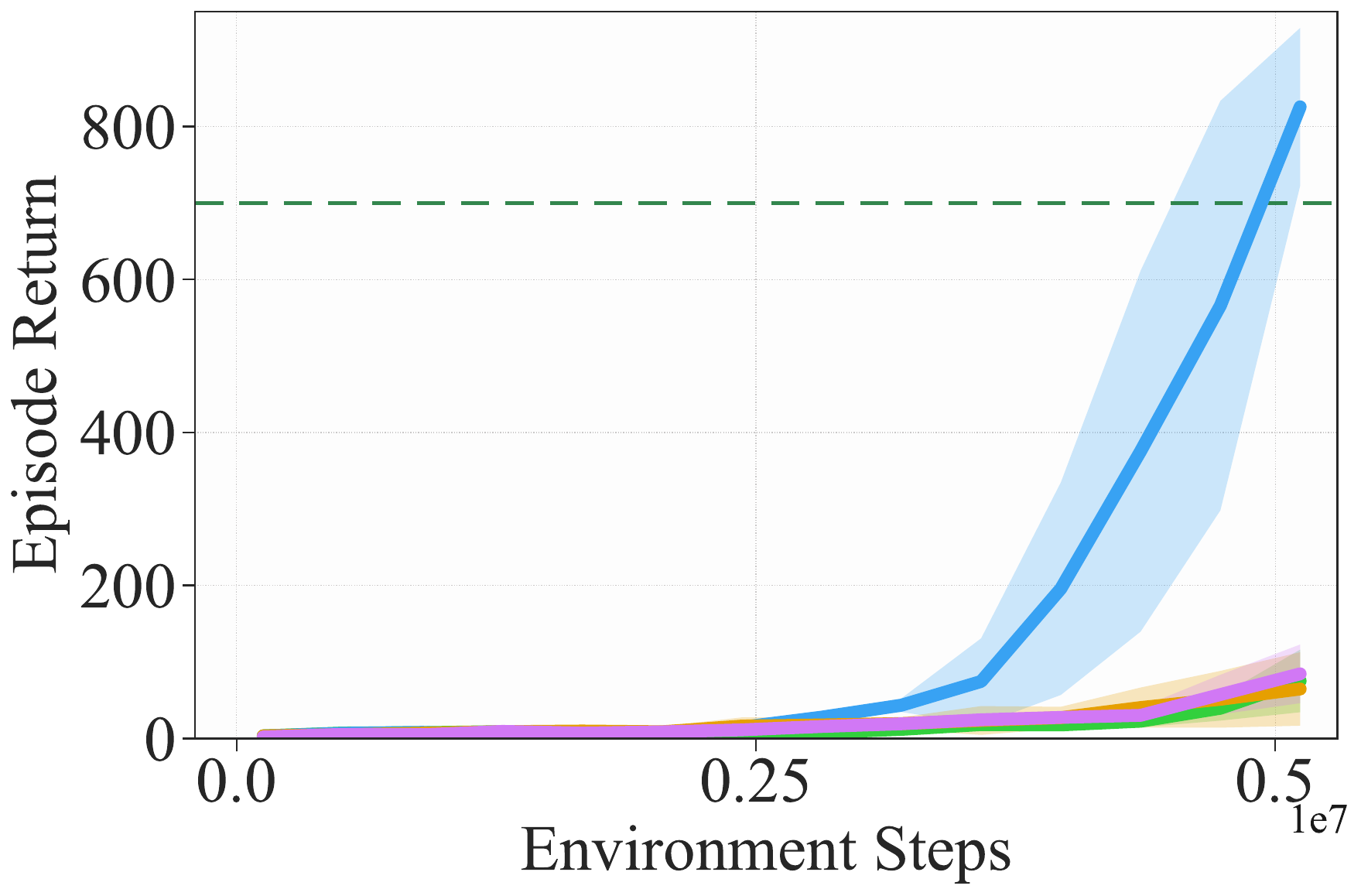}\\
        \vspace{0.01cm}
        \small \hspace{3mm}\textbf{(a)} h1hand\_run\par
    \end{minipage}
    \hspace{0.045\textwidth}
    \begin{minipage}[t]{0.24\textwidth}
        \centering
        \includegraphics[width=\linewidth]{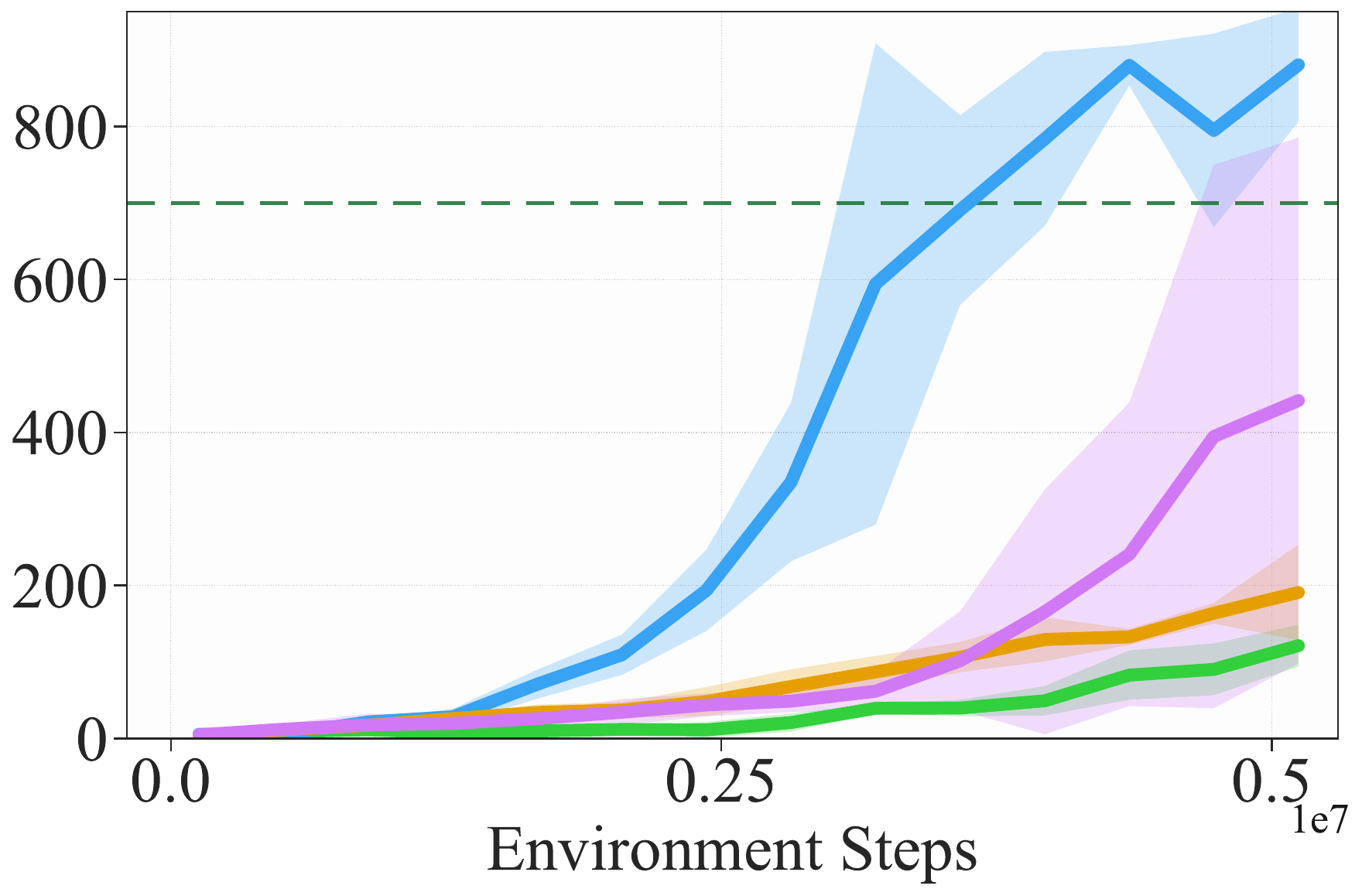}\\
        \vspace{0.01cm}
        \small \hspace{1mm}\textbf{(b)} h1hand\_walk\par
    \end{minipage}
    \hspace{0.045\textwidth}
    \begin{minipage}[t]{0.24\textwidth}
        \centering
        \includegraphics[width=\linewidth]{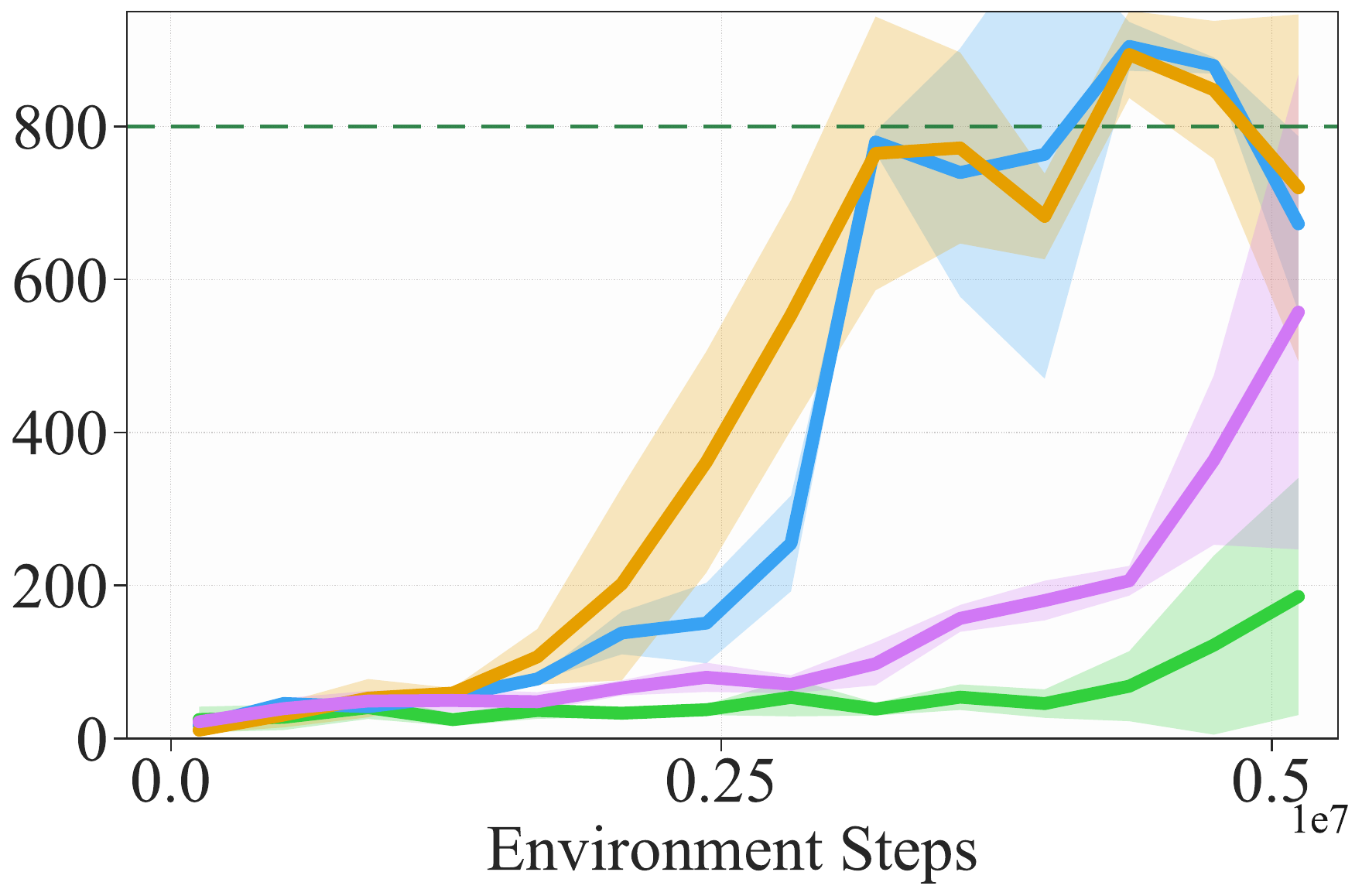}\\
        \vspace{0.01cm}
        \small \hspace{1mm}\textbf{(c)} h1hand\_stand\par
    \end{minipage}

    \vspace{0.3cm}

    \begin{minipage}[t]{0.24\textwidth}
        \centering
        \includegraphics[width=\linewidth]{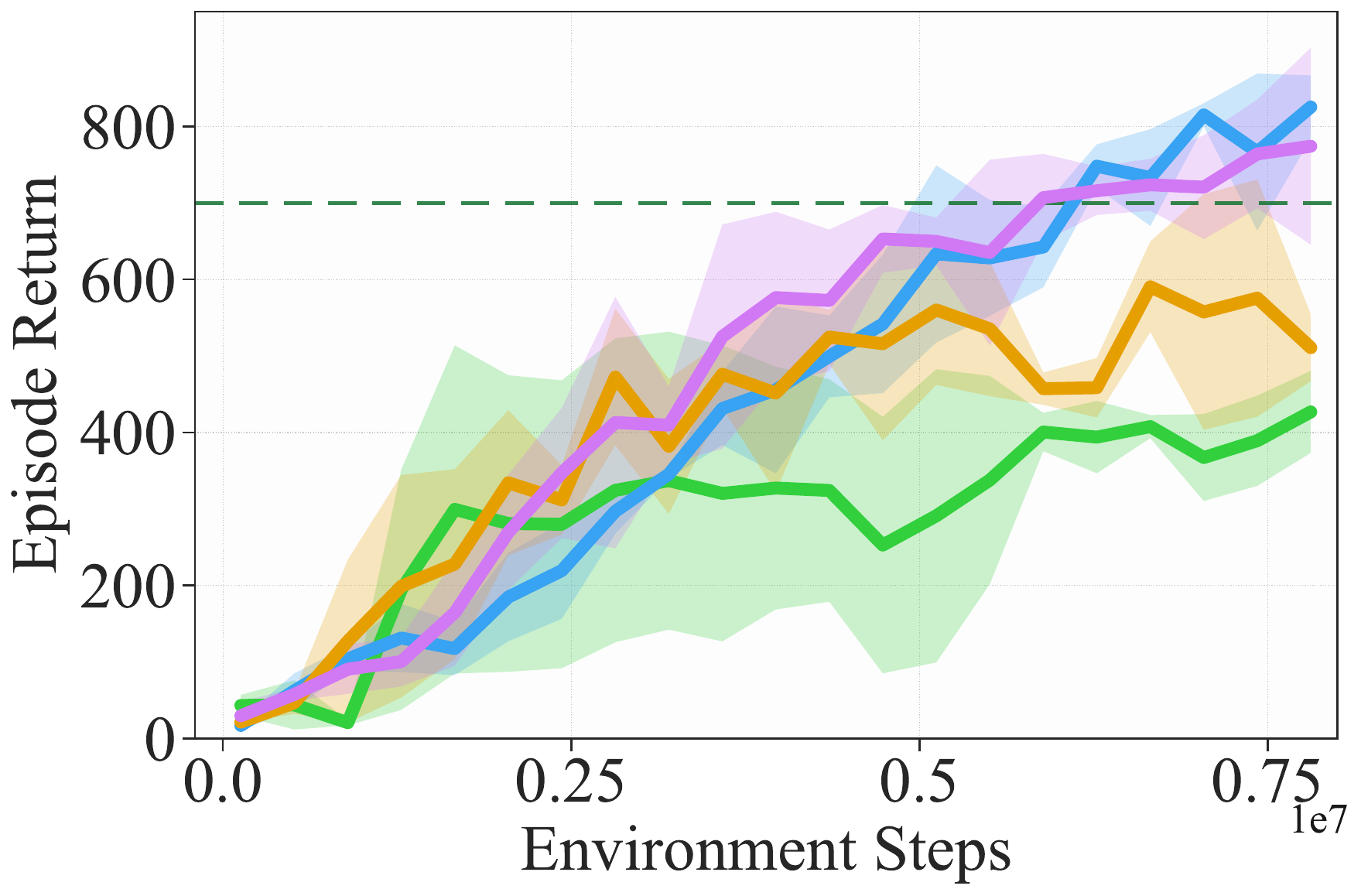}\\
        \vspace{0.01cm}
        \small \hspace{3mm}\textbf{(d)} h1hand\_pole\par
    \end{minipage}
    \hspace{0.045\textwidth}
    \begin{minipage}[t]{0.24\textwidth}
        \centering
        \includegraphics[width=\linewidth]{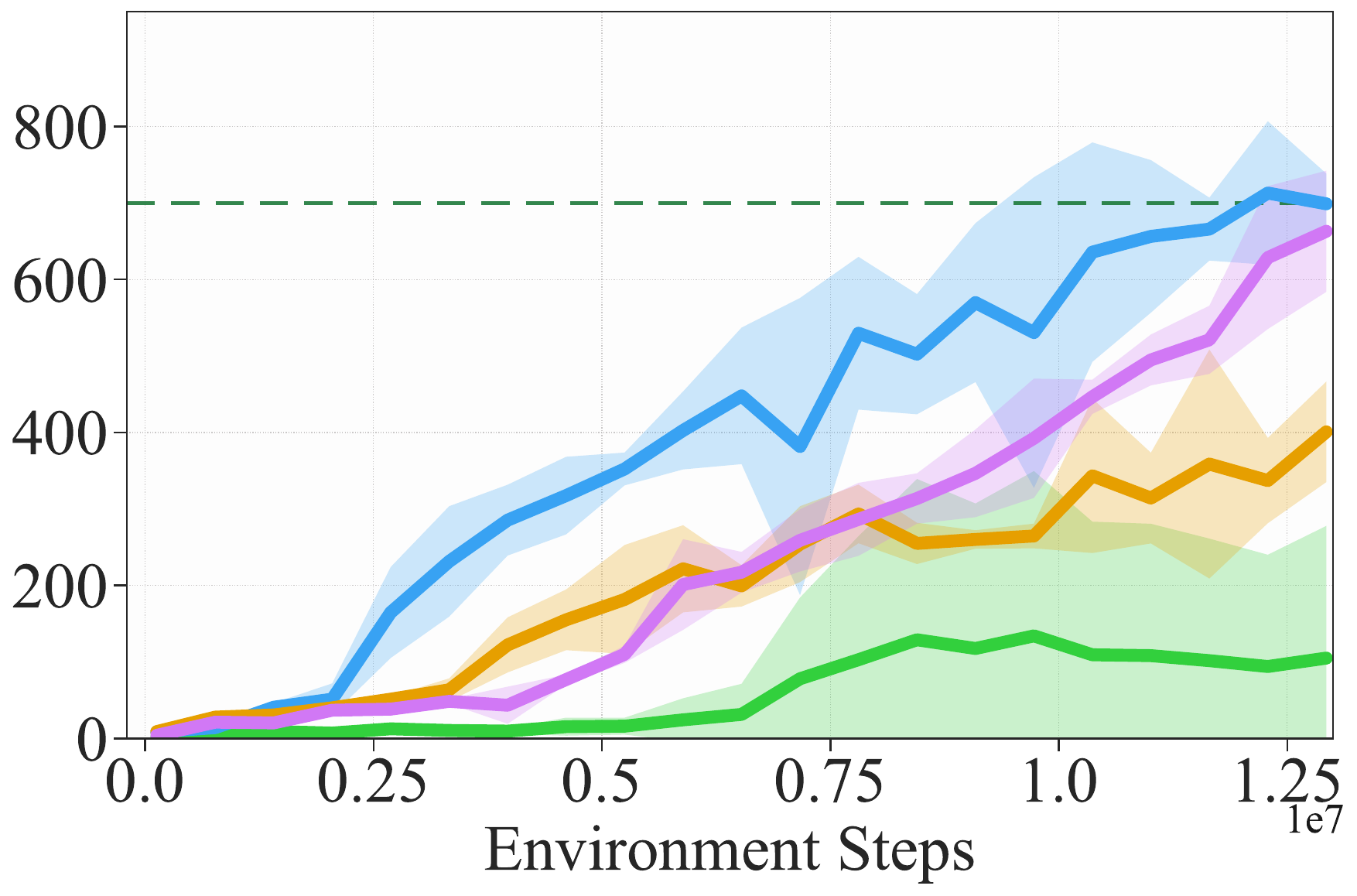}\\
        \vspace{0.01cm}
        \small \hspace{1mm}\textbf{(e)} h1hand\_slide\par
    \end{minipage}
    \hspace{0.045\textwidth}
    \begin{minipage}[t]{0.24\textwidth}
        \centering
        \includegraphics[width=\linewidth]{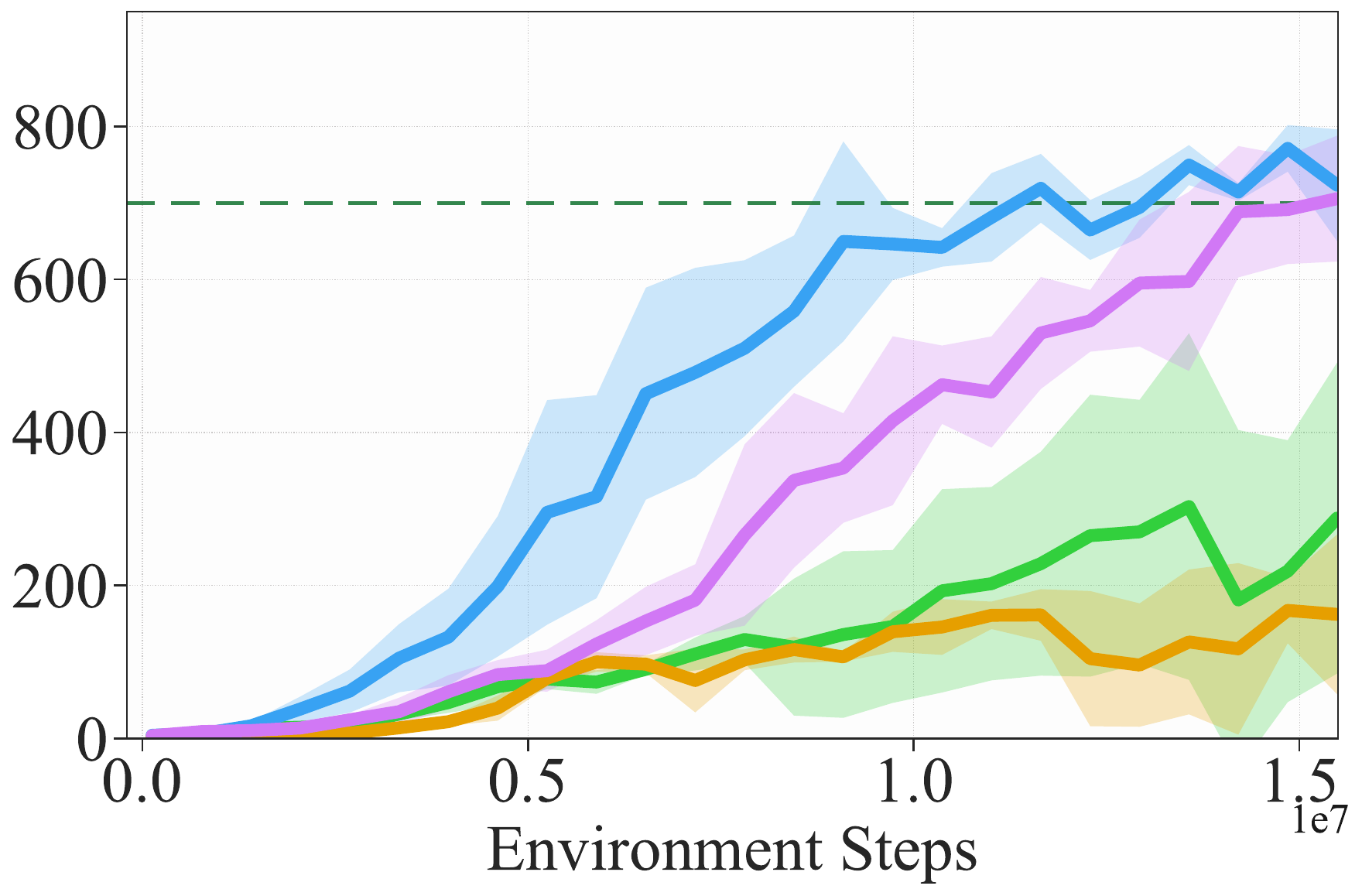}\\
        \vspace{0.01cm}
        \small \hspace{1mm}\textbf{(f)} h1hand\_hurdle\par
    \end{minipage}

    \vspace{0.2cm}

    \begin{minipage}[t]{0.7\textwidth}
        \centering
        \includegraphics[width=\linewidth]{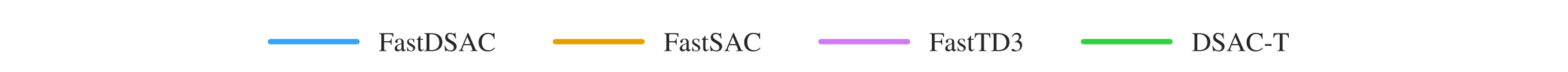}\\
        
    \end{minipage}

    \caption{\textbf{Episode returns on selected HumanoidBench locomotion tasks.}
    Solid lines clearly show the empirical mean over three independent random seeds, and shaded regions consistently indicate one standard deviation.
Dashed horizontal lines mark the corresponding predefined task-specific success thresholds officially used in HumanoidBench.}
    \label{fig:hb_results}
\end{figure*}

\begin{figure*}[t!]
    \centering
    \includegraphics[
        width=\textwidth,
        height=0.18\textheight,
        keepaspectratio
    ]{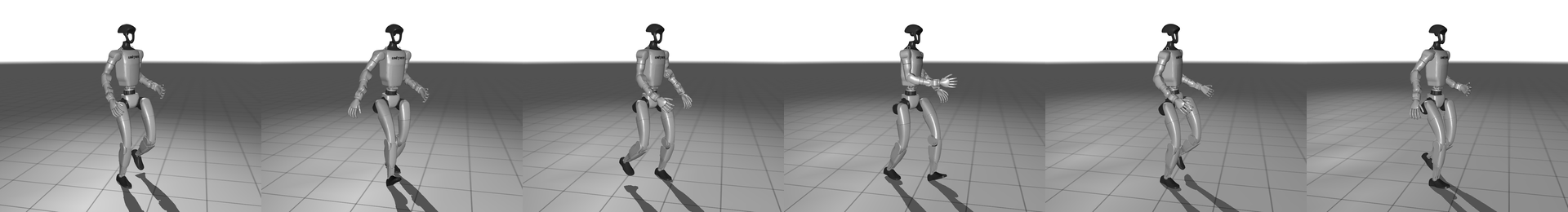}
    \caption{\textbf{Gait visualization under dynamics perturbations.}
    The robot maintains a stable and coordinated walking gait under perturbed mass parameters and ground friction conditions, without obvious posture degradation.}
    \label{fig:gait_robustness}
\end{figure*}

\section{Experiments}
\subsection{Experimental Setups}
The proposed method is evaluated on ten locomotion tasks from two benchmark suites: MuJoCo Playground (4 tasks) and HumanoidBench (6 tasks).
On MuJoCo Playground, we report learning curves as a function of both environment steps as shown in Fig.~\ref{fig:mjstep_results} and wall-clock time as shown in Fig.~\ref{fig:mjclock_results}.
All experiments are run on a single NVIDIA RTX 4090 GPU.
Averaged results over three random seeds are reported, where solid lines and shaded regions denote the mean and standard deviation.
Comparisons are made against FastSAC, FastTD3, and DSAC-T under the same evaluation protocol across both benchmark suites; PPO is additionally included as an on-policy baseline on MuJoCo Playground.

\subsection{Main Results}
\textbf{1) Sample efficiency:}
Fig.~\ref{fig:mjstep_results} reports environment-step learning curves on four MuJoCo Playground locomotion tasks. 
As an on-policy baseline, PPO improves much more slowly and plateaus at markedly lower returns across all tasks, highlighting the advantage of off-policy training in this setting. Among off-policy methods, the baselines exhibit contrasting behaviors. FastSAC rises very quickly at the start of training, but its learning becomes less stable later and shows signs of reduced learning plasticity, which leads to lower final returns. FastTD3 and DSAC-T both exhibit stable learning curves, but their improvement is more gradual, and their final returns remain below FastDSAC under the same environment-step budget. In contrast, FastDSAC combines fast early gains with stable late-stage learning, achieving the highest final return on all four tasks. This advantage is especially evident on the more challenging G1JoystickRoughTerrain task: FastDSAC matches FastSAC in early progress, while continuing to improve without the late-stage regressions observed in FastSAC. The observed behavior supports the effectiveness of the proposed mean-centered truncation for target actions, which mitigates the influence of rare extreme next actions and yields lower-variance TD targets in more complex environments. 

\textbf{2) Time efficiency:}
As shown in Fig.~\ref{fig:mjclock_results}, the learning curves are plotted against wall-clock time. PPO, as an on-policy baseline, improves substantially more slowly in wall-clock time and consistently underperforms the off-policy baselines under the same wall-clock budget.
Among the off-policy baselines, FastSAC shows a higher per-step environment interaction latency, which leads to a longer total wall-clock training time. In contrast, FastTD3, DSAC-T, and FastDSAC are comparable in overall wall-clock training time. More importantly, FastDSAC attains higher returns than FastTD3 earlier in wall-clock time, indicating better time efficiency.
Although FastSAC is slightly faster early on in wall-clock time on the T1Joystick tasks, it reaches an early peak and then declines. FastDSAC surpasses FastSAC shortly after this peak and converges to a higher, more stable return, offering a favorable trade-off for the modest additional wall-clock time. On the G1Joystick tasks, FastDSAC matches FastSAC in real-time learning speed while achieving higher final returns.

\begin{figure}[t!]
    \centering
    \begin{minipage}[t]{0.49\columnwidth}
        \centering
        \includegraphics[width=\linewidth]{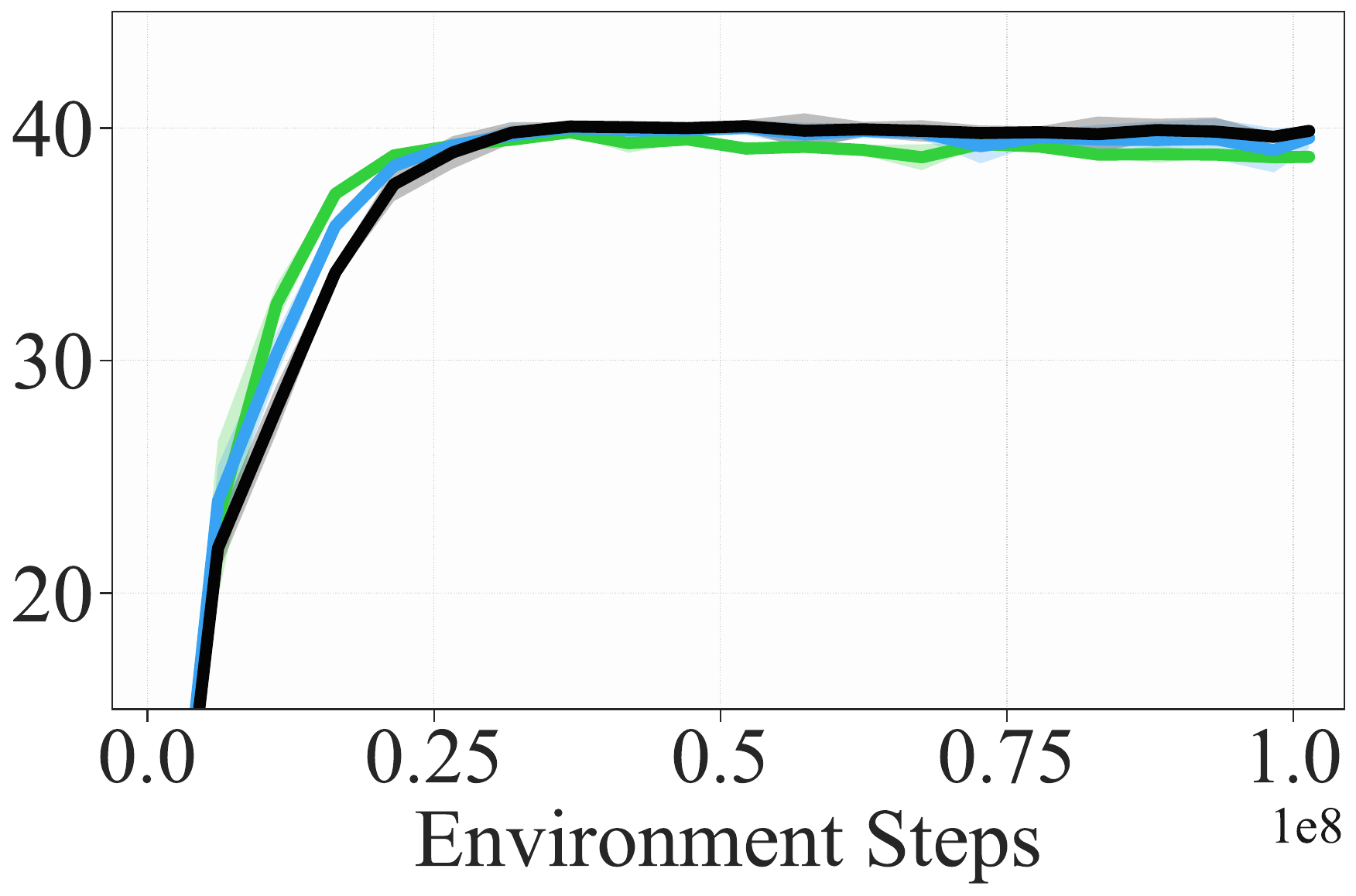}

        \small \textbf{(a)} T1JoystickFlatTerrain\par
    \end{minipage}\hfill
    \begin{minipage}[t]{0.49\columnwidth}
        \centering
        \includegraphics[width=\linewidth]{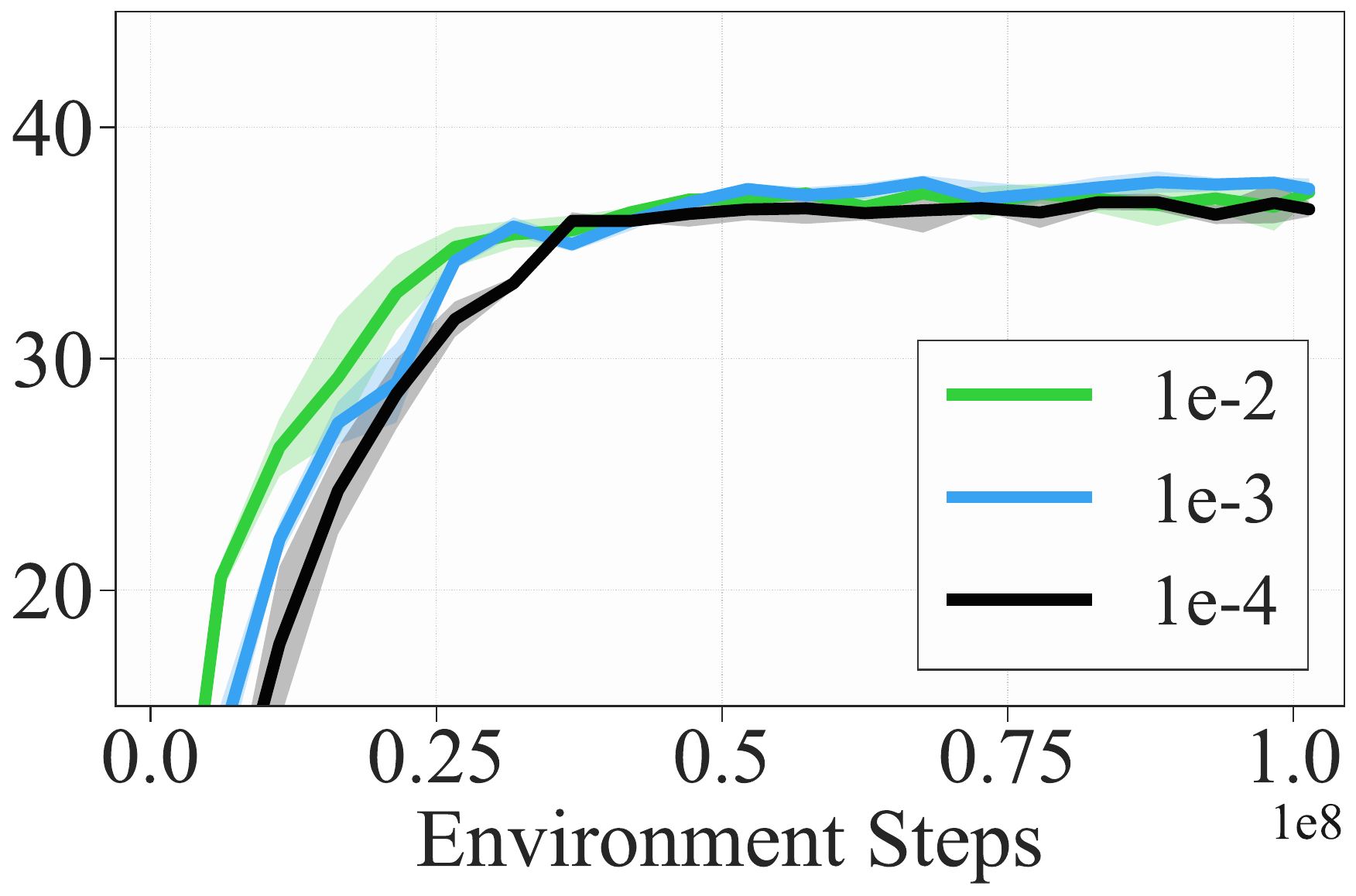}

        \small \textbf{(b)} T1JoystickRoughTerrain\par
    \end{minipage}

    \caption{\textbf{Sensitivity to the truncation radius $c$ in mean-centered target actions.}
    Learning curves on T1JoystickFlatTerrain and T1JoystickRoughTerrain for $c\in\{10^{-2},10^{-3},10^{-4}\}$ in the mean-centered truncation mapping for target actions.
    Solid lines show the mean over three runs, and shaded regions indicate one standard deviation.}
\label{fig:ablation_1}
\end{figure}

\textbf{3) Transition rewards:}
Beyond the episode returns reported in Fig.~\ref{fig:mjstep_results}, transition rewards $r$ in collected tuples $\langle s,a,r,s'\rangle$ are also analyzed on MuJoCo Playground, as shown in Fig.~\ref{fig:mjreward_results}. FastDSAC achieves higher and more stable transition rewards during data collection than FastTD3, FastSAC, DSAC-T, and PPO across all four tasks, indicating that exploration is guided more consistently toward high-reward behaviors and is less affected by rare extremes. Two design choices contribute to this behavior. First, compared with FastTD3 and FastSAC, which use discrete value distributions with fixed intervals, FastDSAC employs a Gaussian distributional critic with adaptively learned variance.
This improves value estimation accuracy and reduces overconfident updates in epistemically uncertain regions, thereby retaining more high-reward transitions during interaction. Second, FastDSAC effectively leverages mean-centered truncation by applying it to the target action in TD error evaluation. This reduces target variance, stabilizes Q-value learning, and makes policy updates more reliable and consistent, leading to improved sampling rewards.
Together, these choices stabilize learning throughout policy optimization and support more stable and higher exploration performance.

\textbf{4) Cross-suite generalization on HumanoidBench:}
Evaluation is further conducted on six HumanoidBench locomotion tasks, as summarized in Fig.~\ref{fig:hb_results}, to examine cross-suite generalization.
Under the same training protocol, FastSAC shows weaker and less consistent outcomes on HumanoidBench than it does on MuJoCo Playground.
In contrast, FastDSAC remains competitive across the HumanoidBench tasks, exhibiting more stable learning behavior across runs and better preserving learning plasticity over training, which enables more reliable improvement under the same protocol.
Across these tasks, FastDSAC is generally on par with FastTD3 or achieves better performance, while both methods substantially outperform DSAC-T by a large margin. Overall, the improvements are not tied to a particular benchmark suite and generalize across evaluation suites.

\textbf{5) Robustness to dynamics perturbations:} To provide an illustrative evaluation under altered system dynamics, we modify the nominal mass parameters and the ground friction coefficient using one representative perturbation setting during forward locomotion, while keeping the remaining simulation and control settings unchanged. As illustrated in Fig.~\ref{fig:gait_robustness}, the robot preserves a stable, natural, and well-coordinated walking gait without obvious posture degradation. This result provides preliminary qualitative evidence that the learned policy can tolerate the considered dynamics perturbation.

\subsection{Ablation Studies}
This subsection presents a series of ablation studies to more systematically evaluate the proposed mean-centered truncation and several related design choices.

\indent\textbf{Impact of the truncation radius $c$.} This ablation examines how FastDSAC depends on the truncation radius $c$ in the mean-centered truncation mapping in Eq.~\eqref{eq:trunc_map}, where $c$ sets the scale of the mean-centered offset and thus controls the allowable deviation of the target action from the policy mean when evaluating target Q-values. As shown in Fig.~\ref{fig:ablation_1}, FastDSAC shows only moderate sensitivity within a practical range overall, and varying $c\in{10^{-2},10^{-3},10^{-4}}$ mainly affects early learning speed while the final returns remain relatively close. This pattern aligns with the role of $c$ in the truncation. Larger $c$ makes the mapping less restrictive and preserves more target action variability, which can speed up early learning, whereas smaller $c$ yields more conservative target actions and can slow early progress. Empirically, $c=10^{-3}$ works well in simpler environments by more strongly suppressing rare extreme target actions, whereas $c=10^{-2}$ performs better in more complex environments by retaining additional variability in the target action to accommodate more complex dynamics. Based on these observations, we use $c=10^{-3}$ as a reasonable and robust default choice. For a new task, we recommend a small logarithmic sweep over ${10^{-2},10^{-3},10^{-4}}$ and selecting $c$ according to both learning speed and training stability. In practice, a larger value can be preferred when learning is overly conservative, while a smaller value can be used when training shows large fluctuations or frequent performance regressions.

\begin{figure}[t!]
    \centering
    \begin{minipage}[t]{0.49\columnwidth}
        \centering
        \includegraphics[width=\linewidth]{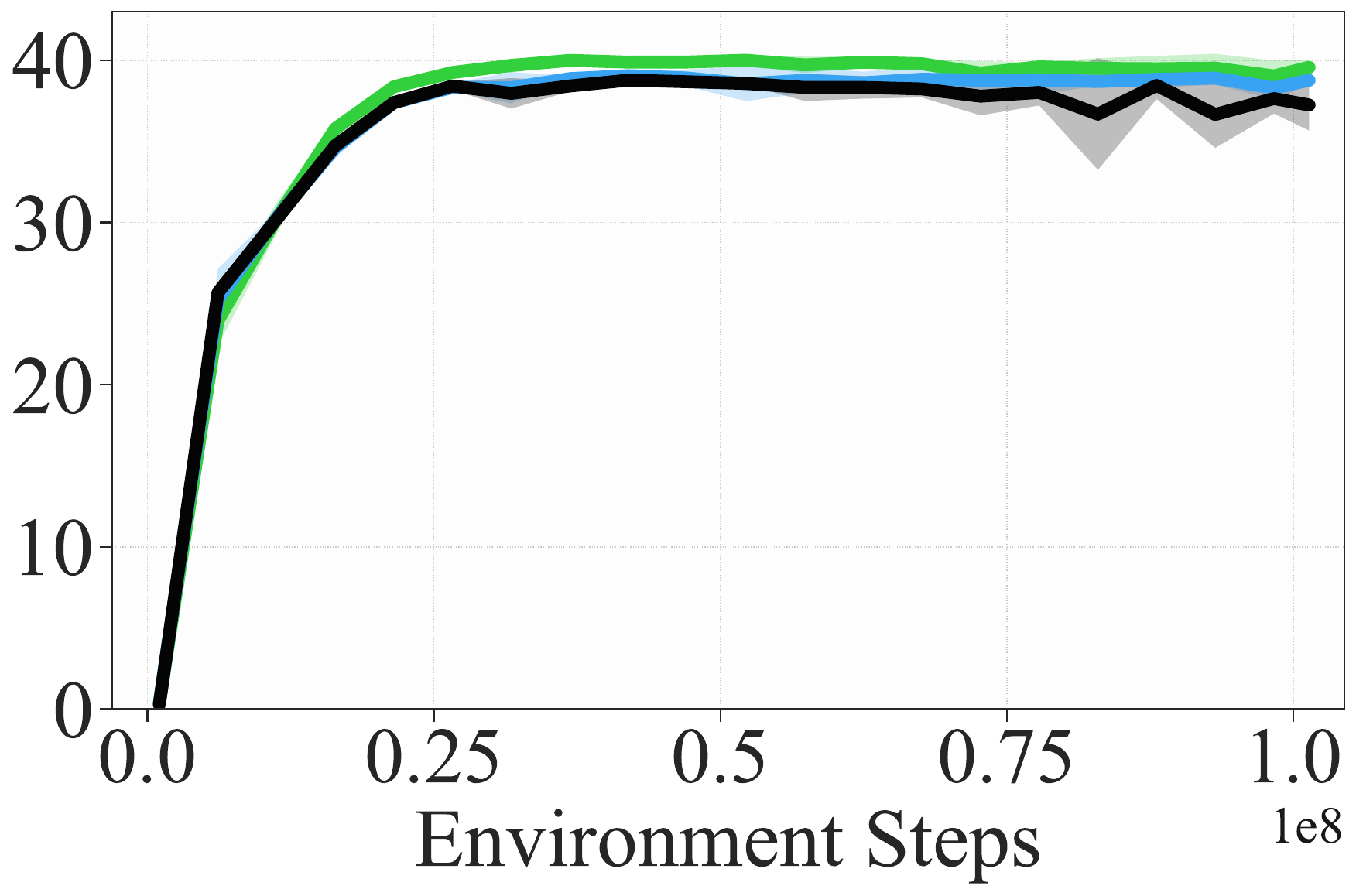}

        \small \textbf{(a)} T1JoystickFlatTerrain\par
    \end{minipage}\hfill
    \begin{minipage}[t]{0.49\columnwidth}
        \centering
        \includegraphics[width=\linewidth]{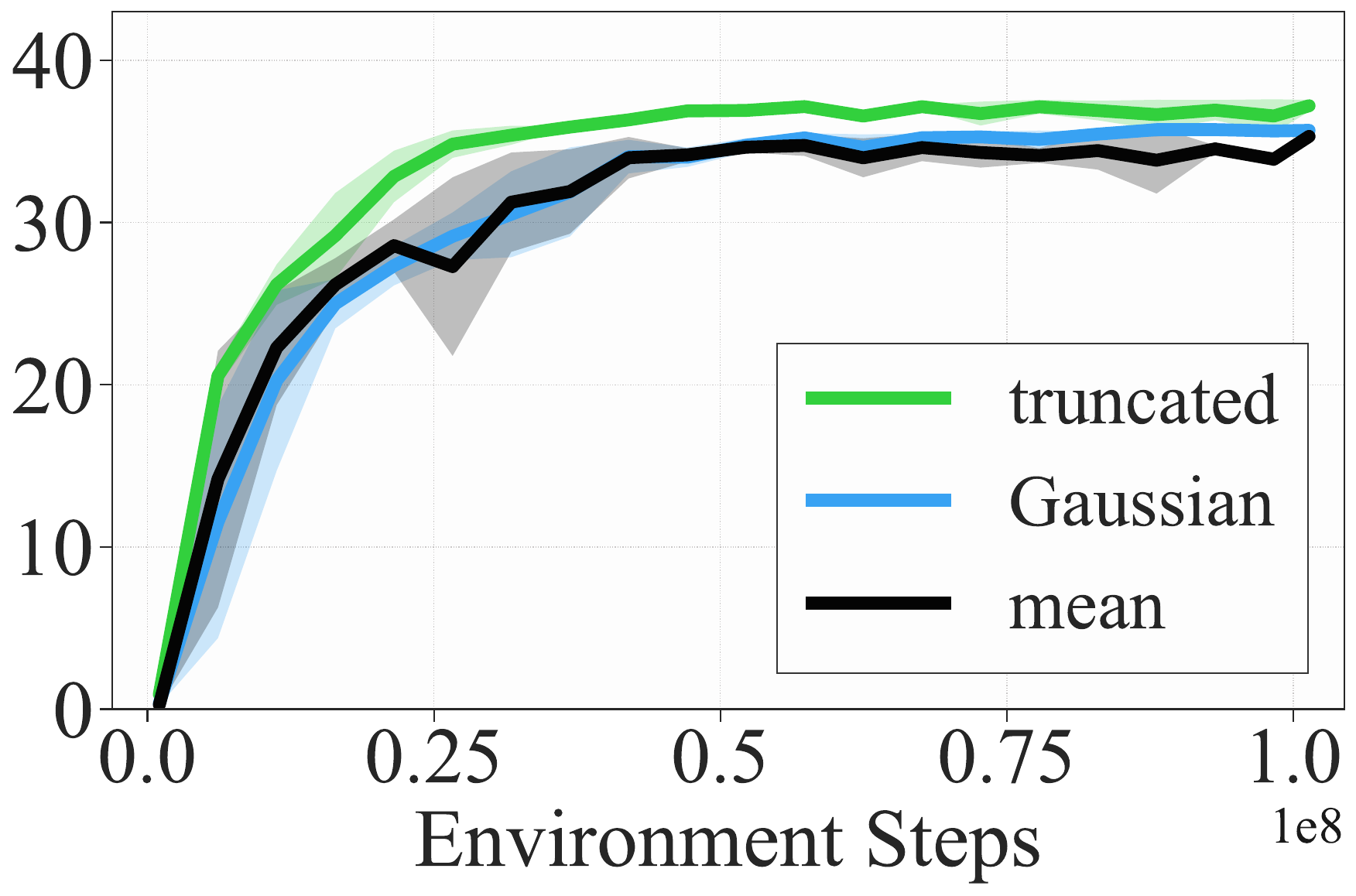}

        \small \textbf{(b)} T1JoystickRoughTerrain\par
    \end{minipage}

    \caption{\textbf{Target action variants in target Q-value evaluation.}
    Learning curves with three choices of the target action used for target Q-value evaluation:
    the target action $a'$ from the mean-centered truncation (\texttt{truncated}),
    the raw Gaussian sample $u'$ without truncation (\texttt{Gaussian}),
    and the deterministic policy mean $\mu_\phi(s')$ (\texttt{mean}).
    Results are shown on T1JoystickFlatTerrain and T1JoystickRoughTerrain.
    Solid lines denote the mean over three runs and shaded regions denote one standard deviation.}
    \label{fig:ablation_2}
\end{figure}

\indent\textbf{Impact of target action choice.}
This ablation evaluates how choices of the target action for target Q-value evaluation influence training dynamics. Besides the target action $a'$ from the mean-centered truncation mapping, two alternatives are considered: using the raw Gaussian sample $u'$ and using the policy mean $\mu_\phi(s')$ as a deterministic target action.
Fig.~\ref{fig:ablation_2} shows that forming the target action via the mean-centered truncation mapping yields the best learning behavior, with only small differences on the simpler flat terrain but more pronounced gaps on the more challenging rough terrain.
Using the Gaussian sample remains competitive, but it often rises more slowly early in training on more complex terrain.
This trend is consistent with the fact that, without the mean-centered truncation, the target action is drawn freely from the Gaussian distribution, which increases the variance of target actions and makes the target Q-values noisier, thereby slowing critic learning.
Using the deterministic mean generally underperforms, as removing stochasticity limits target-action diversity and weakens exploratory coverage, which can lead to lower final returns, most clearly on more complex terrain.

\begin{figure}[t!]
    \centering
    \begin{minipage}[t]{0.49\columnwidth}
        \centering
        \includegraphics[width=\linewidth]{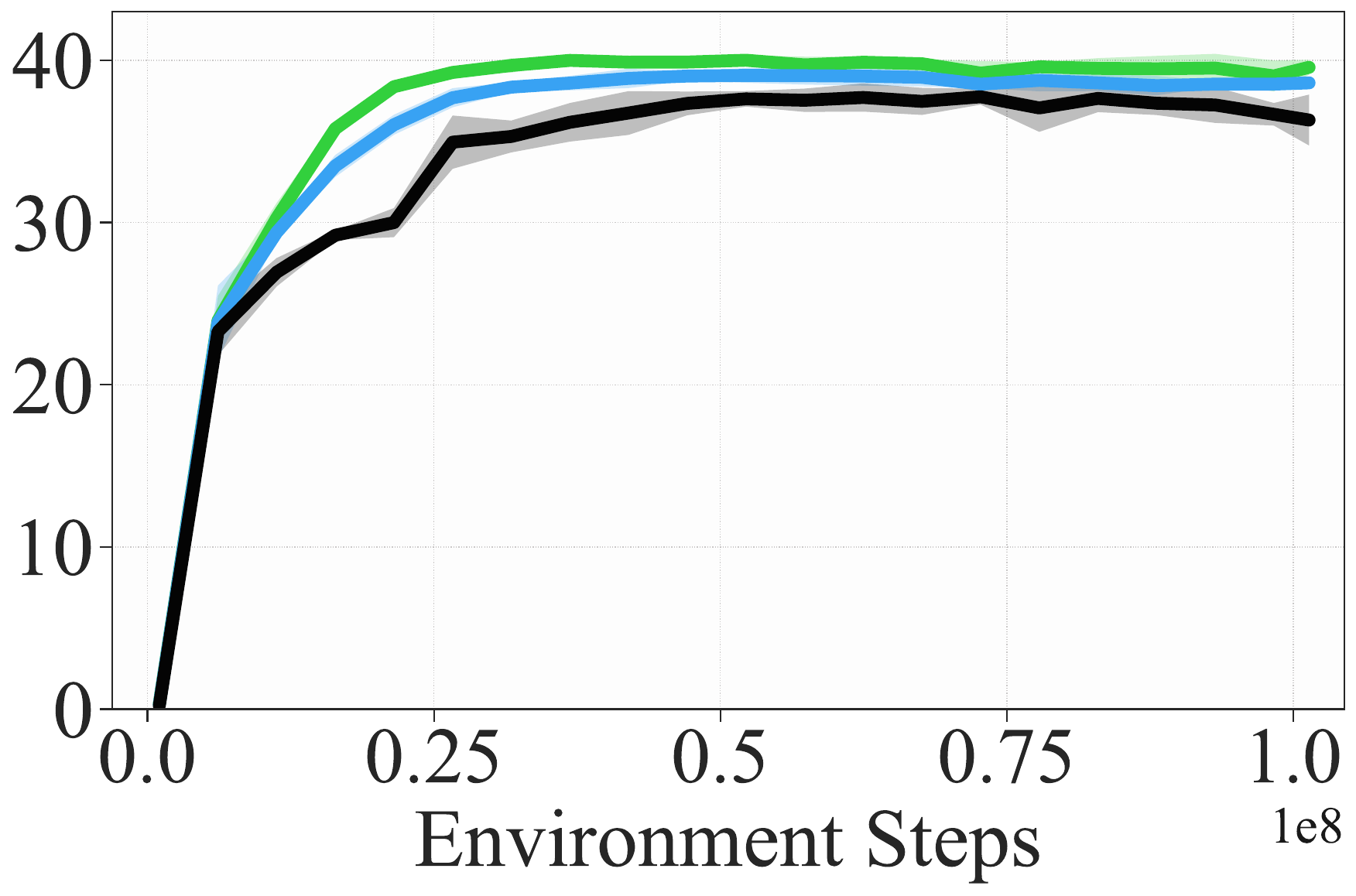}

        \small \textbf{(a)} Scale Env in FlatTerrain\par
    \end{minipage}\hfill
    \begin{minipage}[t]{0.49\columnwidth}
        \centering
        \includegraphics[width=\linewidth]{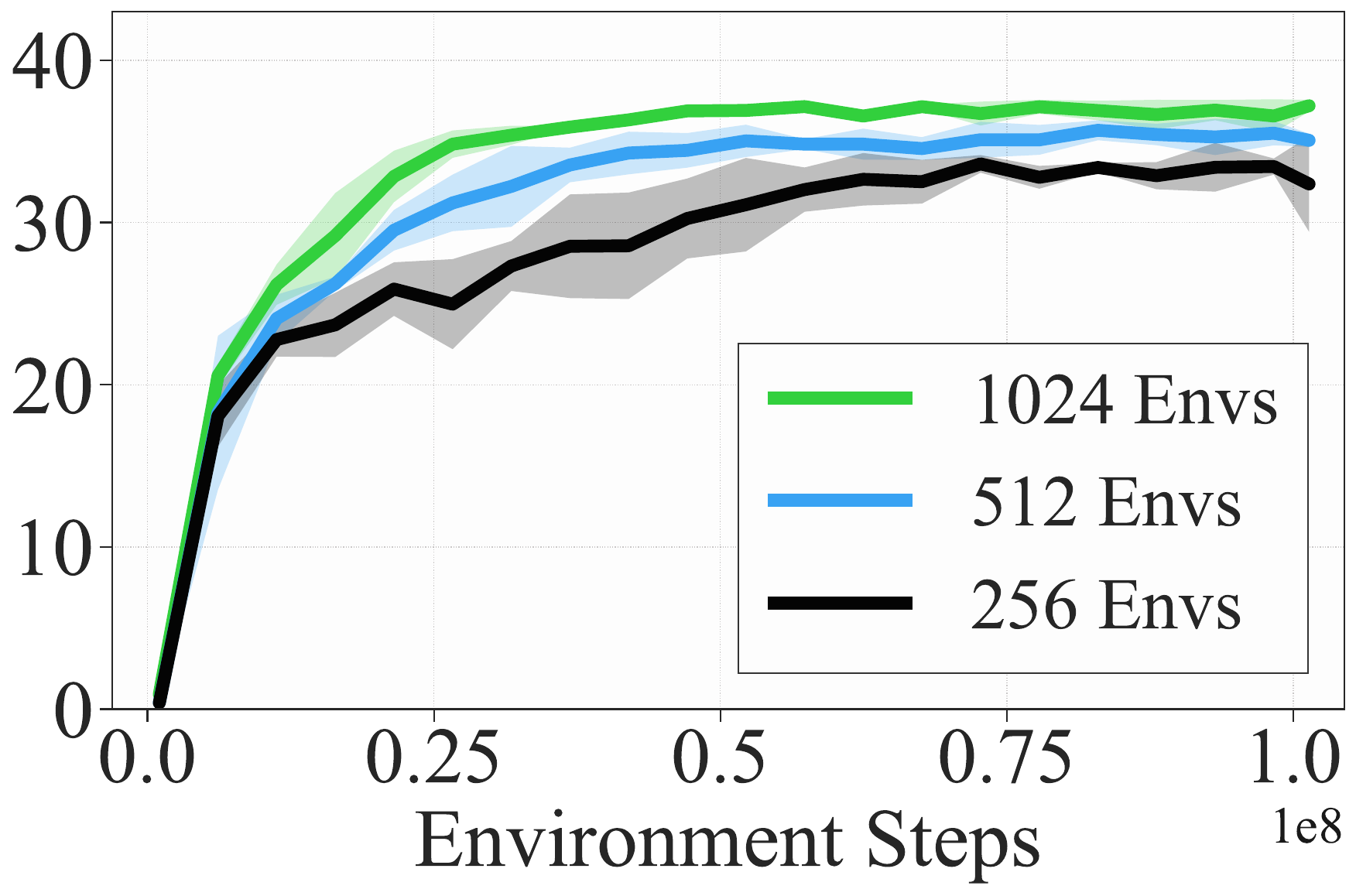}

        \small \textbf{(b)} Scale Env in RoughTerrain\par
    \end{minipage}

    \vspace{0.2cm}
    
    \begin{minipage}[t]{0.49\columnwidth}
        \centering
        \includegraphics[width=\linewidth]{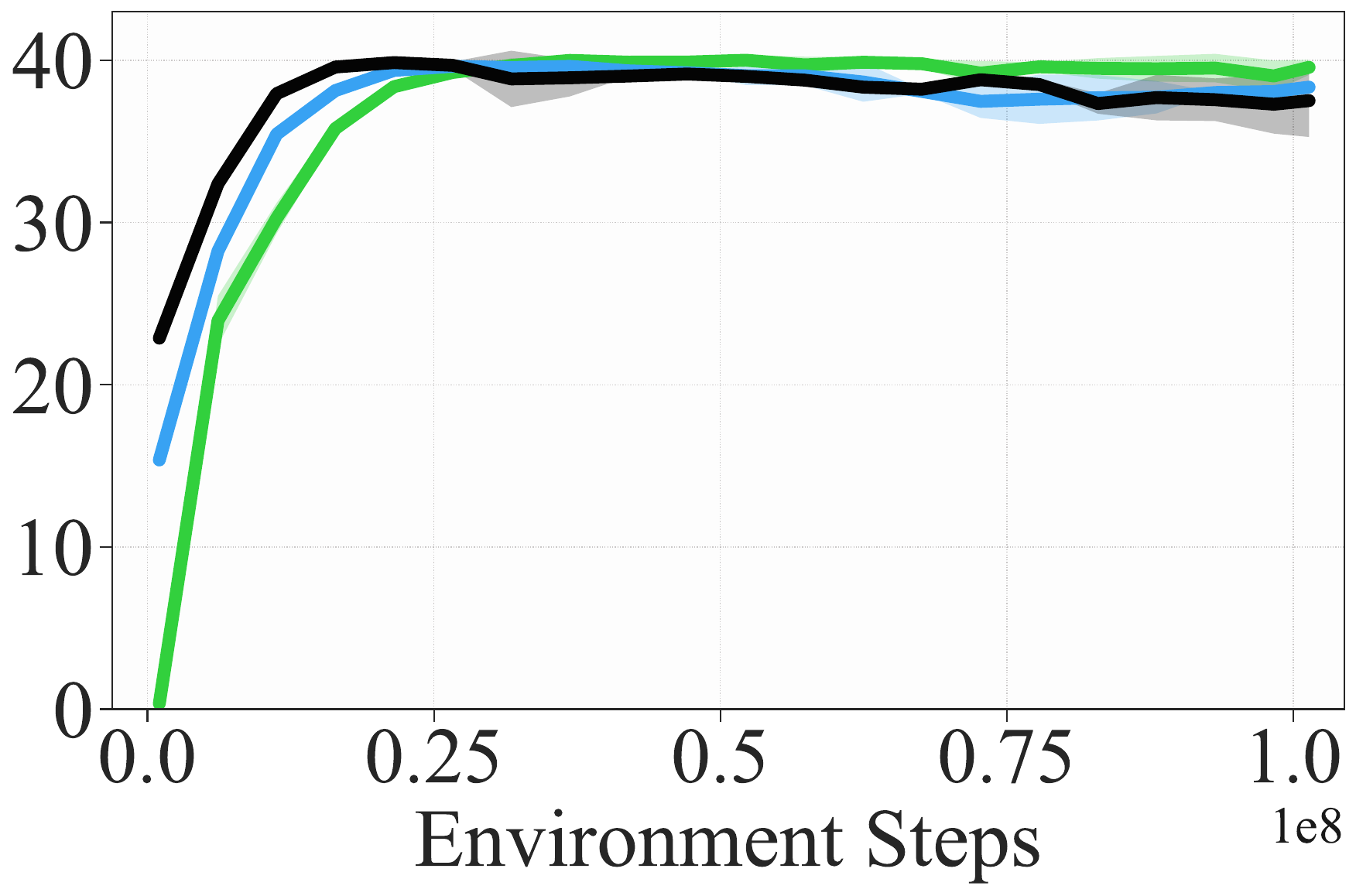}

        \small \textbf{(c)} Scale UTD in FlatTerrain\par
    \end{minipage}\hfill
    \begin{minipage}[t]{0.49\columnwidth}
        \centering
        \includegraphics[width=\linewidth]{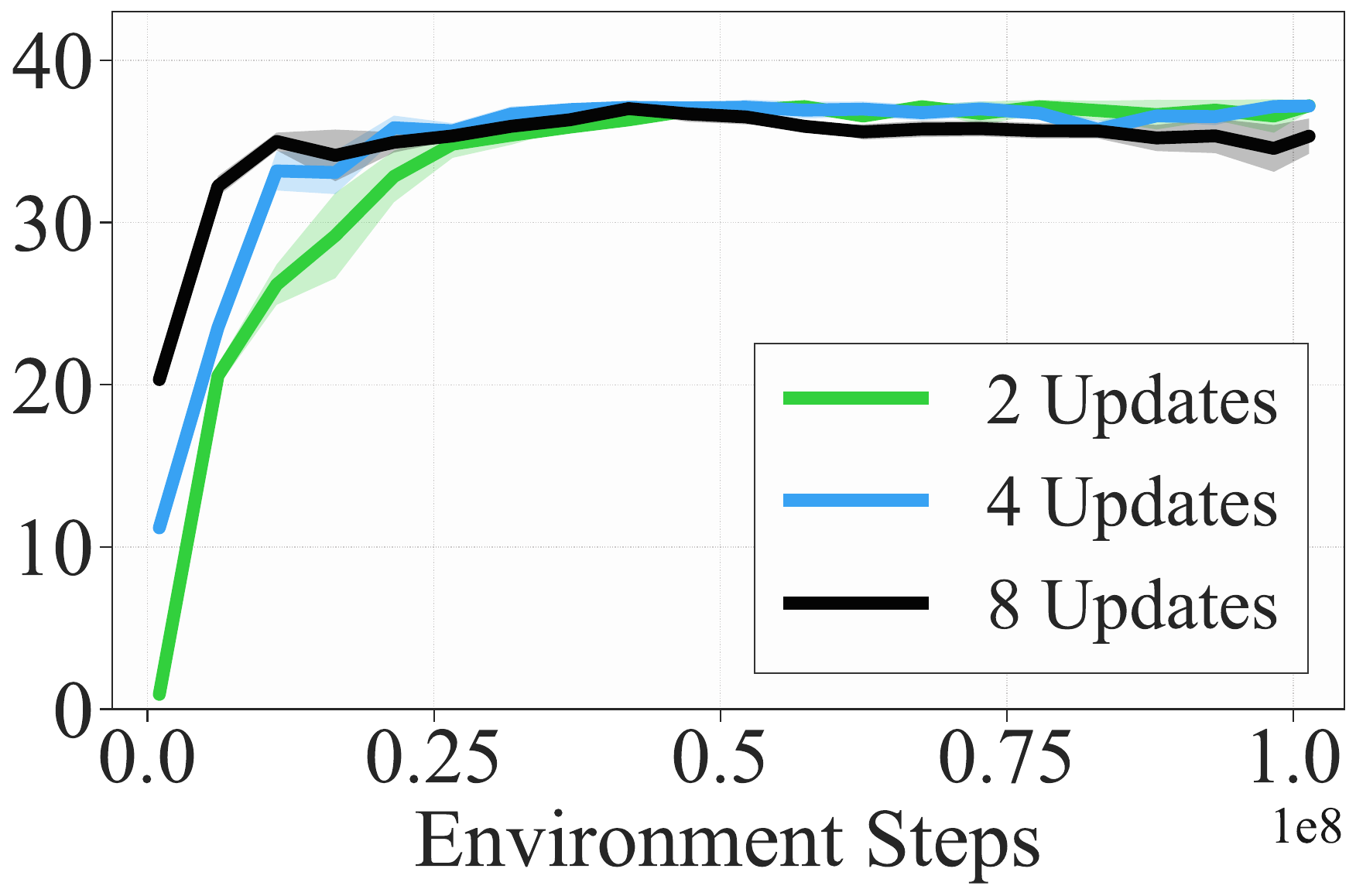}

        \small \textbf{(d)} Scale UTD in RoughTerrain\par
    \end{minipage}
\caption{\textbf{High-throughput scaling with more environments and larger UTD.}
The top row scales the number of parallel environments (256/512/1024), and the bottom row scales the update-to-data ratio (UTD) (2/4/8).
The left column reports T1JoystickFlatTerrain and the right column reports T1JoystickRoughTerrain.
The solid line and shaded region denote the mean and standard deviation over three runs.}
    \label{fig:ablation_3}
\end{figure}

\indent\textbf{Impact of parallelism and UTD ratios on policy plasticity.}
This ablation further evaluates how FastDSAC behaves under higher-throughput settings by scaling parallel data collection and the update-to-data (UTD) ratio. The number of parallel environments is set to $256$, $512$, and $1024$, and the update-to-data (UTD) ratio is set to $2$, $4$, and $8$. As shown in Fig.~\ref{fig:ablation_3}(a) and Fig.~\ref{fig:ablation_3}(b), increasing the number of environments consistently improves performance and does not cause late-stage regressions, indicating that higher parallelism benefits training; moreover, the mean-centered truncation mapping effectively suppresses the additional noise introduced by high-throughput rollouts. In Fig.~\ref{fig:ablation_3}(c) and Fig.~\ref{fig:ablation_3}(d), increasing UTD mainly accelerates early learning, reaching strong returns substantially faster, while converging to a similar final performance. Notably, no clear late-stage decline is observed under larger UTD, which is consistent with the truncation mapping helping preserve learning plasticity throughout training. Overall, FastDSAC performs strongly across these throughput settings.

\begin{figure}[t!]
    \centering
    \begin{minipage}[t]{0.49\columnwidth}
        \centering
        \includegraphics[width=\linewidth]{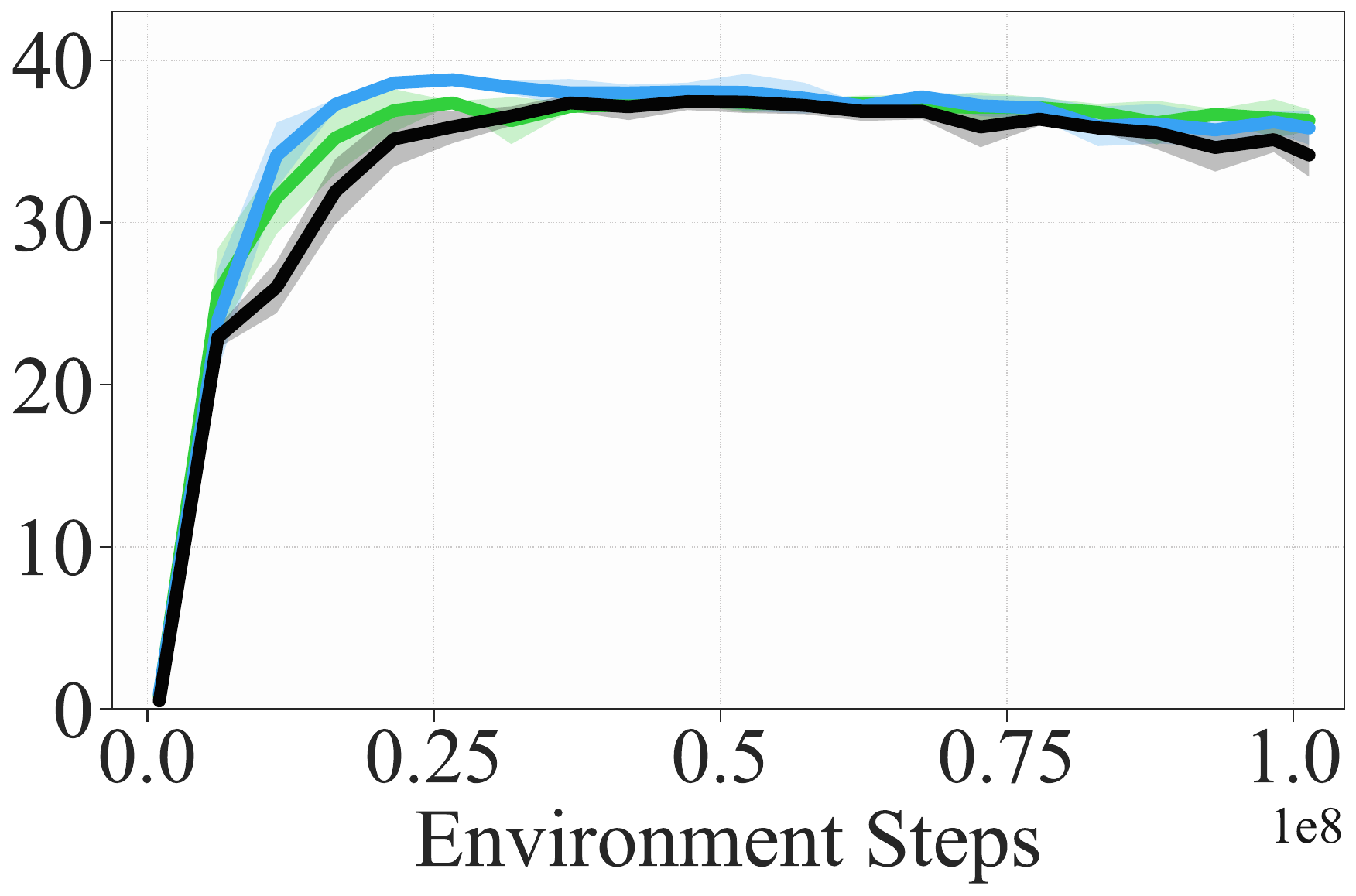}

        \small \textbf{(a)} T1JoystickFlatTerrain\par
    \end{minipage}\hfill
    \begin{minipage}[t]{0.49\columnwidth}
        \centering
        \includegraphics[width=\linewidth]{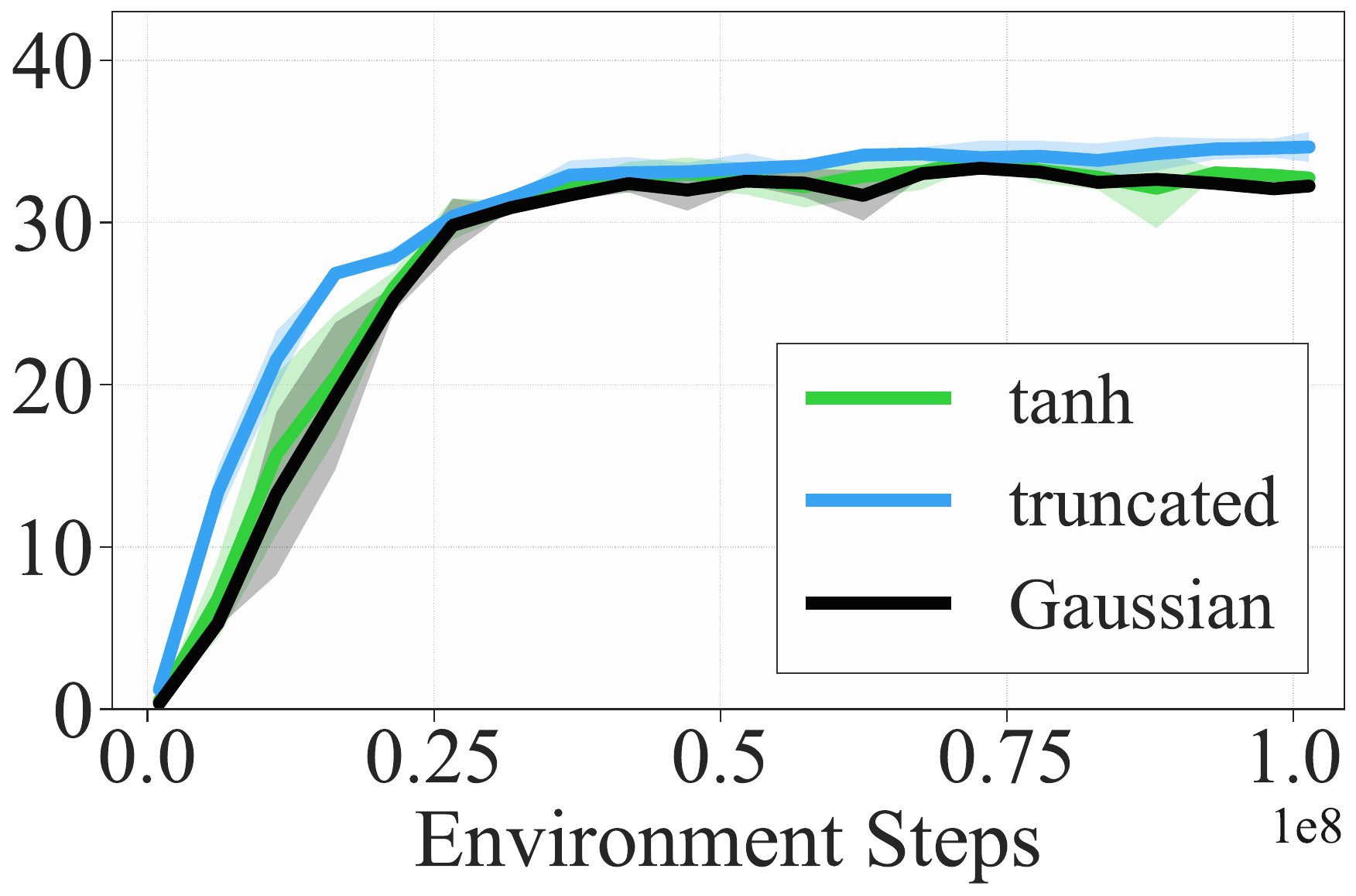}

        \small \textbf{(b)} T1JoystickRoughTerrain\par
    \end{minipage}

    \caption{\textbf{Transferring the mean-centered truncation mapping to SAC.}
    Three SAC variants: raw Gaussian policy (\texttt{Gaussian}), truncated Gaussian policy (\texttt{truncated}), and tanh-squashed Gaussian policy (\texttt{tanh}).
    Results are shown on (a) T1JoystickFlatTerrain and (b) T1JoystickRoughTerrain.
    Solid lines show the mean over three runs, and shaded regions indicate one standard deviation.}
    \label{fig:ablation_4}
\end{figure}

\indent\textbf{Impact of transferring to SAC.}
This final ablation examines whether the mean-centered truncation mapping for target actions is specific to FastDSAC by integrating it into SAC. The mapping is used only to generate the next action for target Q-value computation in the raw Gaussian SAC variant, while all other components remain unchanged. Fig.~\ref{fig:ablation_4} shows that the SAC variant with mean-centered truncation mapping consistently outperforms both the raw Gaussian SAC baseline without truncation and the tanh-squashed SAC baseline, thereby further suggesting that the effect is not specific to FastDSAC in practical applications.

\section{Conclusions}

FastDSAC is a lightweight method for stabilizing entropy-regularized off-policy reinforcement learning under massively parallel sampling and large-batch updates. It applies a mean-centered truncation transformation exclusively to the target action used in the Bellman backup and computes the entropy term using the log-probability induced by the transformed action distribution. By limiting the influence of rare and extreme target actions, FastDSAC improves training stability with negligible computational overhead. Experiments on MuJoCo Playground and HumanoidBench demonstrate that FastDSAC achieves faster and more reliable learning, exhibits fewer late-stage performance regressions, and remains effective at high update-to-data ratios. Ablation studies validate the importance of constraining the target action, show robustness across a practical range of truncation radii, and confirm that the same mechanism also benefits SAC. While these results suggest improved robustness under aggressive optimization, they provide only indirect behavioral evidence of preserved network plasticity. Future work will therefore incorporate plasticity-specific diagnostics, representation-level analyses, qualitative behavior evaluation, and sim-to-real experiments to better characterize the practical benefits and underlying mechanisms of FastDSAC.

\bibliographystyle{IEEEtran}
\bibliography{ref}

\end{document}